\RequirePackage{luatex85}
\documentclass[10pt,twocolumn,letterpaper]{article}

\usepackage{cvpr}              

\usepackage{graphicx}
\usepackage{amsmath}
\usepackage{amssymb}
\usepackage{booktabs}
\usepackage{mathptmx} 
\usepackage{cite}
\usepackage{float}
\usepackage{url}
\usepackage{svg}
\usepackage{pgfplots}
\usepackage{pgfplotstable}
\usepackage{subcaption}
\usepackage{graphicx}
\usepackage{algorithm}
\usepackage{algpseudocode}
\usepackage{etex}
\usepackage{array}

\usepackage[pagebackref,breaklinks,colorlinks]{hyperref}

\usepackage[capitalize]{cleveref}
\crefname{section}{Sec.}{Secs.}
\Crefname{section}{Section}{Sections}
\Crefname{table}{Table}{Tables}
\crefname{table}{Tab.}{Tabs.}

\begin{document}
\setcounter{totalnumber}{50}
\setcounter{topnumber}{50}
\setcounter{bottomnumber}{50}
\renewcommand{\topfraction}{1.0}
\renewcommand{\bottomfraction}{1.0}
\renewcommand{\textfraction}{0.0}
\renewcommand{\floatpagefraction}{1.0}
\title{MCFormer: A Multi-Cost-Volume Network and Comprehensive Benchmark for Particle Image Velocimetry}

\author{Zicheng Lin\\
International School\\
Beijing University of Posts and Telecommunications\\
{\tt\small linzicheng@bupt.edu.cn}
\and
Xiaoqiang Li\\
College of Engineering\\
Peking University\\
{\tt\small xiaoqiangli@pku.edu.cn}
\and
Yichao Wang\\
College of Physics and Optoelectronic Engineering\\
Harbin Engineering University\\
{\tt\small wangyichao@hrbeu.edu.cn}
\and
Chuang Zhu\\
School of Artificial Intelligence\\
Beijing University of Posts and Telecommunications\\
{\tt\small czhu@bupt.edu.cn}
}
\maketitle




\begin{abstract}
Particle Image Velocimetry (PIV) is fundamental to fluid dynamics, yet deep learning applications face significant hurdles. A critical gap exists: the lack of comprehensive evaluation of how diverse optical flow models perform specifically on PIV data, largely due to limitations in available datasets and the absence of a standardized benchmark. This prevents fair comparison and hinders progress. To address this, our primary contribution is a novel, large-scale synthetic PIV benchmark dataset generated from diverse CFD simulations (JHTDB \cite{jhdb} and Blasius). It features unprecedented variety in particle densities, flow velocities, and continuous motion, enabling, for the first time, a standardized and rigorous evaluation of various optical flow and PIV algorithms. Complementing this, we propose Multi Cost Volume PIV (MCFormer), a new deep network architecture leveraging multi-frame temporal information and multiple cost volumes, specifically designed for PIV's sparse nature. Our comprehensive benchmark evaluation, the first of its kind, reveals significant performance variations among adapted optical flow models and demonstrates that MCFormer significantly outperforms existing methods, achieving the lowest overall normalized endpoint error (NEPE). This work provides both a foundational benchmark resource essential for future PIV research and a state-of-the-art method tailored for PIV challenges. We make our benchmark dataset and code publicly available to foster future research in this area.
\end{abstract}

\section{Introduction}
\label{sec:intro}

\begin{figure*}[htb]
    \centering
    \includegraphics[width=0.9\textwidth]{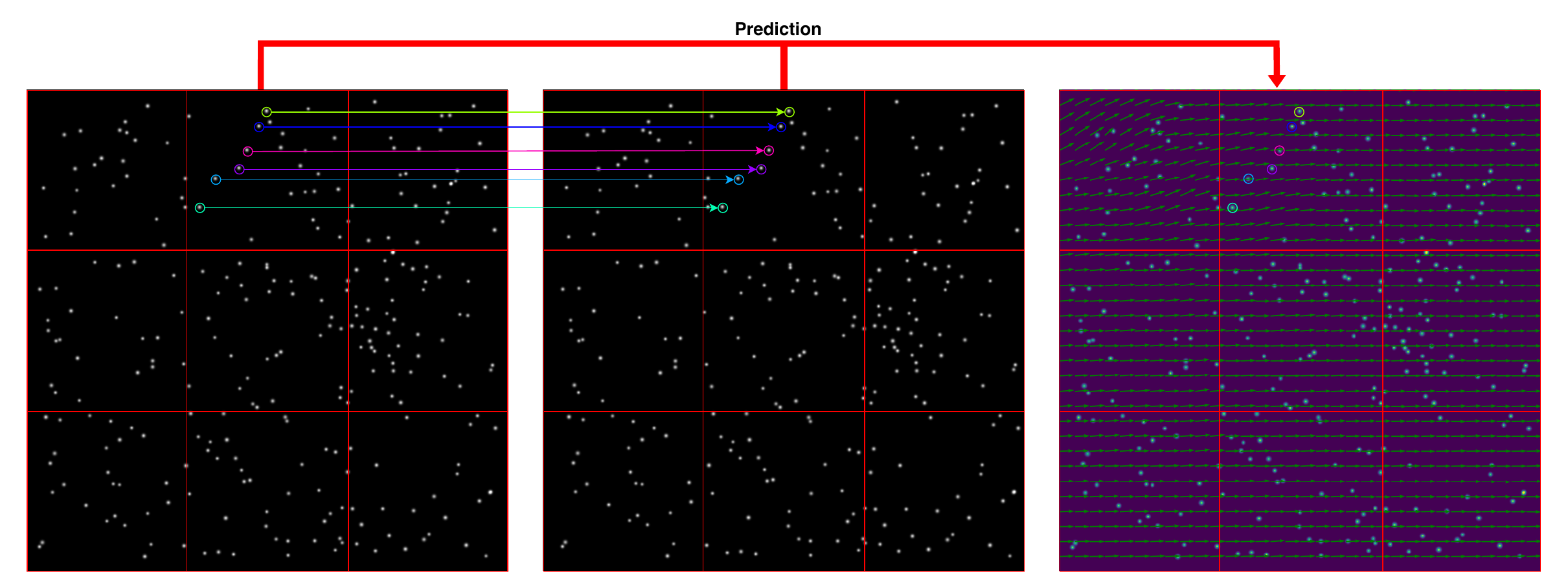}   
    \caption{The principle of Particle Image Velocimetry (PIV)}    
    \label{process} 
\end{figure*}

Particle Image Velocimetry (PIV) is a crucial non-intrusive optical technique in fluid dynamics, enabling the quantification of instantaneous velocity fields within complex flows . By tracking the motion of microscopic tracer particles seeded into the fluid using high-speed cameras, PIV provides invaluable insights into fundamental fluid mechanics across diverse applications like aerodynamics, combustion, and oceanography. The core challenge lies in accurately estimating a dense velocity field from sequences of images that often contain only sparse particle distributions.


Recently, deep learning has emerged as a promising direction for PIV \cite{Lee2017PIVDCNNCD,cai2019dense,litepivnet,TRPIV,pivraft}. However, current deep learning PIV models are often direct adaptations of established optical flow architectures, such as LiteFlowNet \cite{LiteFlowNet}, RAFT \cite{RAFT}, and FlowNetS \cite{flownet}. While these adaptations leverage advancements in optical flow, they may not be inherently optimized for the unique characteristics and challenges of PIV, particularly the prevalent issue of sparse particle distributions. Furthermore, the PIV field currently lacks a comprehensive evaluation of how various optical flow models—ranging from foundational architectures to more recent state-of-the-art designs—perform specifically on diverse PIV data. This gap exists largely because of severe limitations in available PIV datasets. Due to the difficulty of obtaining dense ground truth velocity fields in real-world experiments, most PIV datasets rely on Computational Fluid Dynamics (CFD) simulations \cite{Lee2017PIVDCNNCD,cai2019dense,litepivnet,TRPIV}. Existing synthetic datasets often lack diversity, primarily featuring high particle densities \cite{cai2019dense,Lee2017PIVDCNNCD,TRPIV}, limited velocity ranges \cite{litepivnet}, or discrepancies compared to real fluid dynamics. Compounding this is the critical lack of a widely accepted, standardized benchmark dataset \cite{Lee2017PIVDCNNCD,cai2019dense,litepivnet,TRPIV}. This absence makes it exceedingly difficult to perform fair and rigorous comparisons between different algorithms or to systematically assess the suitability of various optical flow backbones for the PIV task. Models are often trained and evaluated on custom datasets \cite{Lee2017PIVDCNNCD,cai2019dense,litepivnet,TRPIV}, hindering reproducible research and impeding progress. Consequently, the true potential and limitations of applying modern optical flow techniques to PIV remain incompletely understood, and models trained on non-standardized data often exhibit poor generalization.

Moreover, another critical limitation inherent in most current deep learning PIV methods \cite{Lee2017PIVDCNNCD,cai2019dense,litepivnet,pivraft}, often inherited from their optical flow origins, is their reliance on only two consecutive frames for velocity estimation. Fluid motion is inherently a continuous temporal process. By discarding information from preceding or succeeding frames, these methods fail to leverage rich temporal context that could significantly improve accuracy, especially for resolving sub-pixel displacements and capturing complex, time-varying flow structures. While some attempts have been made to incorporate temporal information, such as recurrently using the previous flow prediction \cite{TRPIV}, these approaches may not fully capture longer-range temporal dependencies and have shown limited or inconsistent improvements \cite{TRPIV}.

To address these critical gaps—the lack of a diverse benchmark, the absence of comprehensive model evaluation, the reliance on adapted architectures, and the underutilization of temporal information—we make the following contributions:

A New Comprehensive PIV Benchmark Dataset: We introduce a large-scale synthetic PIV dataset generated from diverse CFD simulations (JHTDB \cite{jhdb}). Designed as a standardized benchmark, it features unprecedented diversity in particle densities, flow velocities, and continuous motion, enabling robust training and, crucially, fair evaluation and comparison of various models, including different optical flow backbones, on PIV data.

Multi Cost Volume PIV (MC-PIV) Network: To better leverage temporal context, moving beyond simple two-frame approaches, we introduce the Multi Cost Volume PIV (MC-PIV) Network. While some existing multi-frame optical flow architectures often focus on strategies like recurrently refining a single predicted flow field or passing features warped by previous flow estimates through the network—effectively optimizing based on evolving flow fields—MC-PIV pioneers an alternative strategy tailored for PIV. It effectively leverages multi-frame temporal information, employing attention mechanisms, and crucially, explores the optimization based on explicitly constructing and integrating information from multiple, distinct cost volumes. This approach allows MC-PIV to capture richer spatio-temporal dynamics and infer high-fidelity flow fields, offering a novel pathway for tackling PIV's unique challenges, particularly in sparse and complex flow conditions.

Comprehensive Benchmark Evaluation: We provide the first comprehensive benchmark results for a wide range of existing optical flow and PIV models on our diverse dataset, establishing baseline performances and revealing their strengths and weaknesses in the context of PIV.

Our extensive experiments demonstrate that MC-PIV significantly outperforms existing methods across all challenging conditions on the new benchmark. The evaluation also underscores the performance variations among different adapted optical flow models when applied to PIV, highlighting the need for specialized approaches and rigorous benchmarking, facilitated by our proposed dataset.


\begin{figure*}[htbp]
    \centering
    \includegraphics[width=1\textwidth]{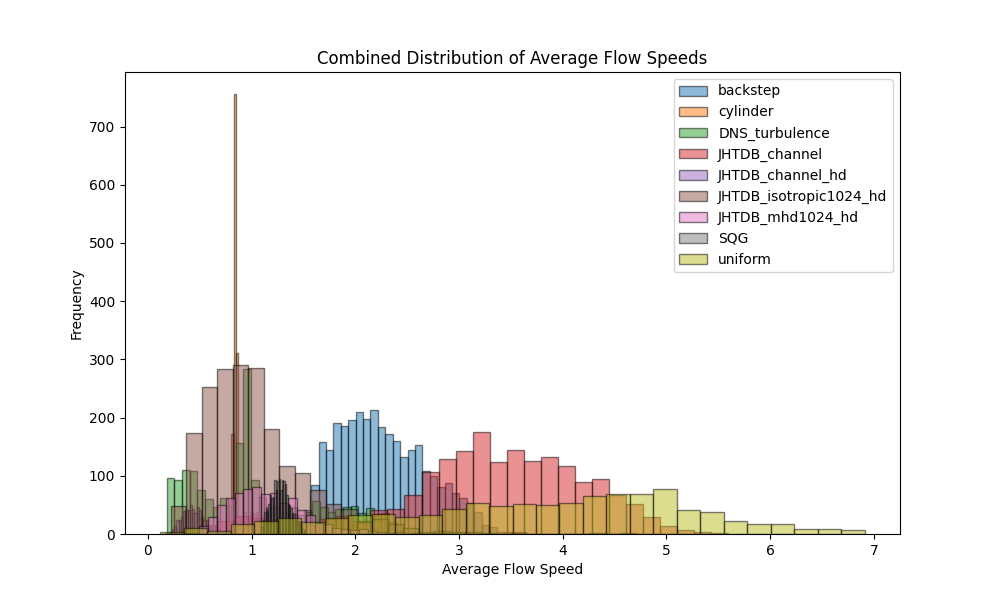}   
    \caption{The Flow Distribution of Cai's PIV Dataset\cite{cai2019dense}}    
    \label{Table:1} 
\end{figure*}
\section{Related Work}

\subsection{Deep Learning PIV Algorithms:} 
    
Several deep learning algorithms have been proposed for PIV. Lee et al. (2017) introduced PIV-DCNN \cite{Lee2017PIVDCNNCD}, which applies cascaded convolutional networks to extract features from images and calculate the cost volume map at different levels for each pixel. The network then aggregates the vectors from different levels to obtain the final result. This network effectively mirrors the WIDIM \cite{WIDIM} approach, replacing the feature extraction component with a convolutional network.

With the advancement of deep learning, researchers have explored its potential for learning the relationships between fluid fields at different time steps. One such attempt is the Time-Resolved Particle Image Velocimetry Algorithm \cite{TRPIV}. The backbone of TR-PIV is LiteFlowNet \cite{LiteFlowNet}, a U-Net-like architecture that uses convolution as its basic building block. LiteFlowNet consists of two main components: a feature extraction part (NetC) and an optical flow estimation part (NetE). The output resolution of this network is half that of the input. While LiteFlowNet is designed for optical flow estimation from a single pair of images, TR-PIV incorporates an additional layer in NetE, increasing the output resolution to match the input resolution. To enable multi-frame PIV analysis, TR-PIV processes the output flow field from the previous time step with the first image feature of the current time step and uses it as input to NetE along with the second image feature.

This method of incorporating temporal information makes the network operate similarly to a recurrent network. However, TR-PIV does not exhibit a significant advantage over LiteFlowNet on the same dataset and even performs worse on some datasets. This raises an intriguing question: why does incorporating more information lead to decreased performance? We hypothesize that this might be due to the datasets used for training. The datasets primarily used for training TR-PIV have high particle densities, providing sufficient data for the model to determine the flow field. Adding potentially irrelevant data might confuse the model.

More recently, Lagemann et al. (2021) adapted the Recurrent All-Pairs Field Transforms (RAFT) architecture \cite{pivraft} for PIV analysis, introducing RAFT-PIV \cite{pivraft}. Unlike the coarse-to-fine refinement strategy common in FlowNet/LiteFlowNet architectures, RAFT operates differently. It extracts features from both images using a shared encoder, computes correlations between all pairs of feature vectors to build a 4D correlation volume (and subsequently a pyramid by pooling), and then iteratively updates a high-resolution flow field using a convolutional Gated Recurrent Unit (Conv-GRU) that looks up values from the correlation volume. They specifically proposed \textbf{RAFT32-PIV}, a variant designed to work directly on smaller image patches (32x32 pixels) without spatial downsampling during feature extraction. This approach allows the model to maintain high-resolution details throughout the process. RAFT32-PIV demonstrated state-of-the-art performance on several benchmark PIV datasets, showing high accuracy and the ability to resolve fine flow structures, outperforming both traditional methods and prior deep learning approaches like PIV-LiteFlowNet under many conditions, especially when dealing with more realistic (less idealized) particle image characteristics. The iterative refinement mechanism appears key to its strong performance.

\subsection{PIV dataset:}


\begin{figure}[htb]
    \centering
    \begin{subfigure}[b]{0.35\textwidth}
        \centering
        \includegraphics[width=\textwidth]{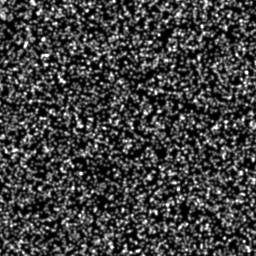}
        \label{}
    \end{subfigure}
    \hfill
    \begin{subfigure}[b]{0.35\textwidth}
        \centering
        \includegraphics[width=\textwidth]{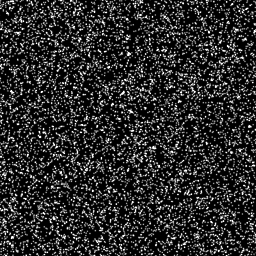}
        \label{}
    \end{subfigure}
    \hfill
    \begin{subfigure}[b]{0.35\textwidth}
        \centering
        \includegraphics[width=\textwidth]{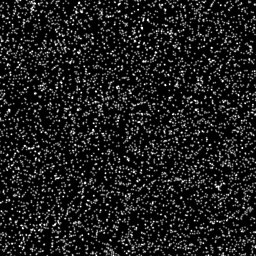}
    \end{subfigure}
    \caption{The examples of the Cai's Dataset\cite{cai2019dense}}
    \label{cai_dataset}
\end{figure}

Lee et al. (2017) introduced PIV-DCNN \cite{Lee2017PIVDCNNCD}, a deep learning approach for PIV, using a synthetic dataset for training and evaluation. This dataset comprises simulated particle image pairs generated using polynomial-based flow fields and two methods: a Particle Image Generator (PIG) and image warping from real PIV images. Gaussian white noise was added for robustness. In Lee's dataset, the particle density was fixed at 0.05, and datasets were generated with varying flow field velocities. PIV-DCNN exhibited superior predictive performance when the magnitude of the flow field displacement was an integer value. However, it is important to note that the majority of the flow fields were not derived from complex fluid dynamics models. This simplification might introduce discrepancies between the dataset and real-world flow fields, potentially hindering the model's ability to accurately predict flow fields based on the motion of surrounding particles in more complex scenarios.

Guo et al. \cite{TRPIV} constructed a synthetic dataset for training and evaluating deep learning models for Time-Resolved PIV. The dataset comprises 12,000 groups of multi-frame particle images and corresponding ground truth velocity fields, simulating various flow patterns, including uniform flow, backward-facing step flow, and DNS turbulence. Each group consists of three-frame images with adaptively transformed velocity fields to mimic the slight variations observed in real TR-PIV experiments. However, this dataset lacks particle density diversity, and only the uniform fluid field exhibits speed diversity.

Cai et al. \cite{cai2019dense} generated a synthetic PIV dataset for training deep learning models (see examples in Figure \ref{density}). This dataset consists of over 10,000 pairs of particle images and corresponding ground-truth velocity fields, representing diverse flow patterns, including uniform flow, backward-stepping flow, vortex shedding over a circular cylinder, 2D turbulent flow (DNS-turbulence), and a surface quasi-geostrophic (SQG) model of sea flow. The particle images were generated with varying parameters, such as particle diameter, seeding density, and peak intensity, to ensure diversity and robustness during training. The motion fields were obtained from computational fluid dynamics (CFD) simulations and publicly available datasets like the Johns Hopkins Turbulence Databases (JHTDB) \cite{jhdb}, covering various flow conditions and Reynolds numbers. This comprehensive dataset enables the training of deep learning models capable of accurately and efficiently estimating fluid motion from particle image pairs in various flow scenarios. However, the datasets are discontinuous. Moreover, when generating datasets with different flow velocities, they did not scale the flow velocity of a large-scale simulation dataset like the JHTDB. Instead, they directly generated uniform backstep cylinder datasets with different flow velocities using a small CFD flow field generator, resulting in an uneven distribution of flow fields with varying fluid velocities.

\section{A Comprehensive PIV Benchmark Dataset}

To address the limitations of existing Particle Image Velocimetry (PIV) datasets – namely their lack of diversity in particle density and flow dynamics, reliance on non-standardized data, and difficulty in facilitating fair model comparisons\cite{Lee2017PIVDCNNCD,cai2019dense,litepivnet,TRPIV,pivraft} – we introduce a new, large-scale synthetic PIV benchmark dataset. This dataset serves as our primary contribution, providing a standardized and challenging platform for developing and evaluating PIV algorithms. 
\subsection{The Proposed PIV Benchmark Dataset}
Our benchmark is built upon high-fidelity ground truth velocity fields derived from both Computational Fluid Dynamics (CFD) simulations and fundamental flow solutions, ensuring physical realism. We utilize five diverse flow types to capture a broad spectrum of fluid dynamics phenomena:

Homogeneous Buoyancy Driven Turbulence (Mixing) \cite{mixing}: Simulates the mixing of two miscible fluids, driven by buoyancy forces (from JHTDB \cite{mixing}).

Forced MHD Turbulence (MHD) \cite{mhd}: Incompressible magnetohydrodynamic turbulence driven by Taylor-Green forcing (from JHTDB \cite{mhd}).

Forced Isotropic Turbulence (Isotropic) \cite{isotropic}: Statistically stationary forced isotropic turbulence (from JHTDB \cite{isotropic}).

Turbulent Channel Flow (Channel) \cite{channel}: Direct numerical simulation of turbulent flow between two parallel plates (from JHTDB \cite{channel}).

Laminar Boundary Layer Flow (Blasius): In addition to the complex turbulent flows from JHTDB, we include a fifth flow type representing a laminar boundary layer. This field is generated from the numerical solution of the Blasius equation, which describes the steady, two-dimensional flow over a semi-infinite flat plate. 

These flows provide a wide range of dynamic conditions. To further enhance diversity, specific scaling strategies were applied (detailed in Section 3.2), and particle images were generated across three distinct particle densities: dense (0.01 ppp), moderate (0.0025 ppp), and sparse (0.001 ppp). The detailed methodology for generating the particle images from these flow fields is described next. Furthermore, data augmentation techniques (noise, occlusion, glare) were applied during training to enhance model robustness.

\subsection{Picture Generation:}
This section details the methodology used to generate synthetic particle image pairs from the ground truth velocity fields described in Section 3.1.

\textbf{Flow Field Scaling:}
To generate diverse flow speeds for the four turbulent JHTDB datasets, we scaled the fluid fields. Directly amplifying velocity can distort physics, so analogous to optical magnification, we extracted regions of $\frac{1}{4}$ and $\frac{1}{8}$  the size of the original fluid field and upsampled them back using interpolation. Fluid velocities within these regions were then amplified by factors of 4 and 8, respectively. First-order Lagrangian interpolation was used for 4x, and fourth-order for 8x magnification. This generates 1x, 4x, and 8x relative velocity conditions for the turbulent flows. The Blasius flow utilized a fixed 50x amplification factor without this sub-region extraction process.

\textbf{Particle Motion and Rendering:}
We assume particles perfectly follow the local fluid velocity vector. The position $(x_1, y_1)$ of a particle at time $t_1$ is calculated from its position $(x_0, y_0)$ at $t_0$ and the local flow vector $(f_x, f_y)$ at $(x_0, y_0)$:
$$(x_1, y_1) = (x_0 + f_x, y_0 + f_y).$$
Particles are rendered onto the image canvas using a Gaussian intensity profile, following the PIV-DCNN formula [13]:
    \[I(x,y) = I_0 \exp \left[ \frac{-(x-x_0)^2-(y-y_0)^2}{(1/8) d_p^2} \right],\]
where $d_p$ is the particle diameter located at $(x_0, y_0)$, and $I_0$ is the peak intensity. 

\begin{figure}
    \includegraphics[width=0.5\textwidth]{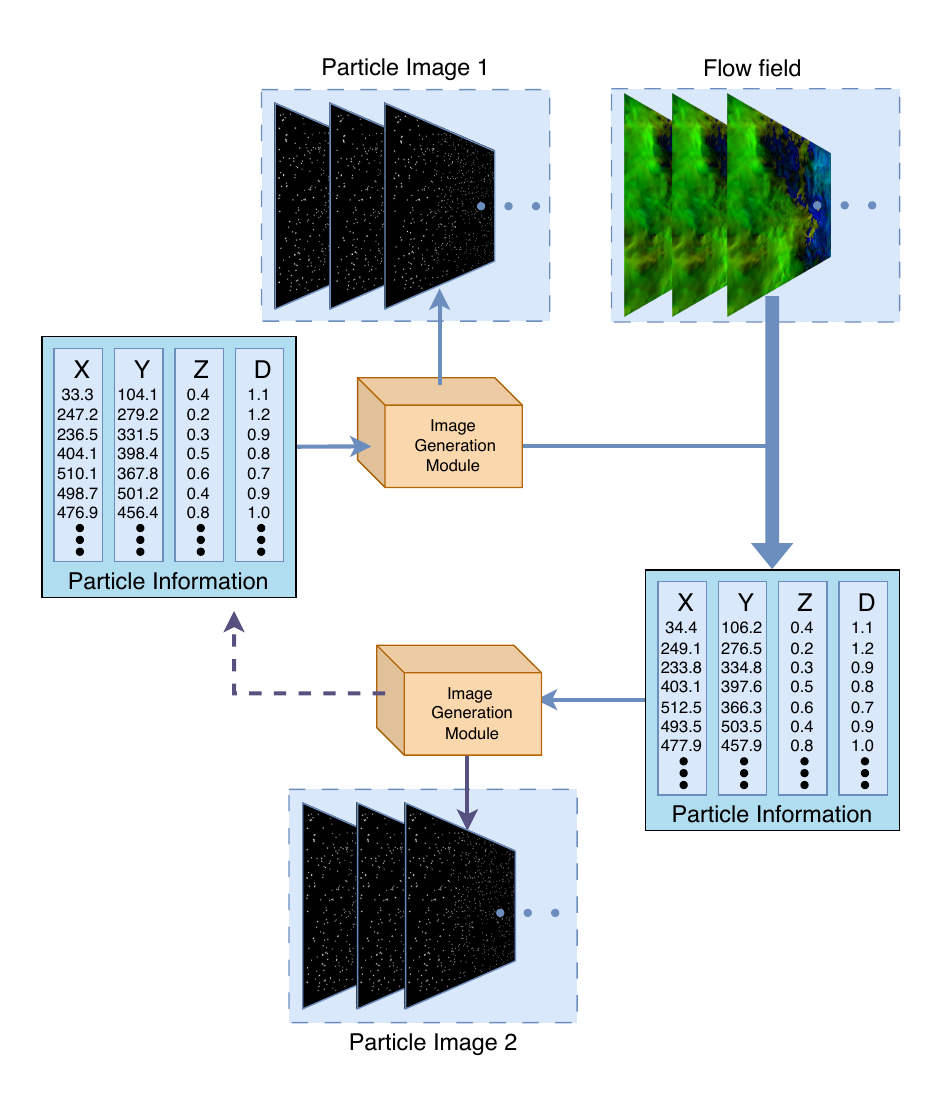}   
    \caption{Flowchart of the iterative Particle Image Generation Process. The module uses particle info at $t_0$ to create Image 1, calculates positions at $t_1$ , removes outliers, creates Image 2, adds new particles, and outputs updated info for the next cycle. Particle Image 2 becomes Image 1 of the subsequent iteration. }    
    \label{PIG:1} 
\end{figure}

\textbf{Iterative Generation Module(Figure \ref{PIG:1}) }
Our particle image generator is illustrated in Figure \ref{PIG:1}.

The module takes the flow field and current particle information as input. It outputs a pair of particle images (Image 1 at $t_0$ , Image 2 at $t_1$) and the updated particle information for the next time step ($t_1$).Generate Image 1 at time $t_0$based on input particle info. Calculate displaced particle positions at $t_1$ using the flow field and motion equation. Remove particles that moved outside image boundaries. Generate Image 2 at time $t_1$ using the updated positions of remaining particles. Calculate and add new particles to maintain density (details below). Return the image pair and the final particle list at $t_1$.



\subsection{The Description of Synthetic PIV Datasets} 
Our comprehensive synthetic PIV benchmark dataset is built upon the five diverse fluid dynamics types detailed in Section 3.1. To ensure variability, we generated particle images across three distinct density levels: dense (0.01 particles/pixel), moderate (0.0025 particles/pixel), and sparse (0.001 particles/pixel). Furthermore, for the four turbulent flow types derived from JHTDB (Mixing, MHD, Isotropic, Channel), we applied three velocity scaling factors (1x, 4x, and 8x) as described in Section 3.2, creating nine distinct dynamic conditions for each of these turbulent flows. The fifth flow type, the laminar Blasius boundary layer, utilized a fixed amplification factor during its generation (see Section 3.2) and was therefore included only at its base scaling configuration across the three densities.

Consequently, the dataset comprises a total of (4 turbulent flows × 3 densities × 3 scalings) + (1 laminar flow × 3 densities × 1 scaling) = 39 unique experimental conditions. For each condition, we generated 500 sequential image pairs, capturing different time instances of the flow. This culminates in a total benchmark size of 19,500 image pairs. Following standard practice, we designated 70\% of these pairs (13,650) for model training and reserved the remaining 30\% (5,850) for testing and evaluation. Table 1 provides a summary of the velocity distribution characteristics across the different generated fluid dynamics data.

\begin{table}[ht]
\centering
\caption{Flow field velocity's Mean and Standard Deviation}
\label{velTable}
\begin{tabular}{@{}lccc@{}}
\toprule
Dataset & Scaling & Mean & Std \\ \midrule
Channel & 1 & 0.8939 & 0.0843 \\
Channel & 4 & 3.5547 & 0.2675 \\
Channel & 8 & 7.0262 & 0.5067 \\
Isotropic1024 & 1 & 1.0388 & 0.4291 \\
Isotropic1024 & 4 & 4.5454 & 1.8380 \\
Isotropic1024 & 8 & 8.6357 & 4.2400 \\
MHD1024 & 1 & 0.2393 & 0.1406 \\
MHD1024 & 4 & 1.0064 & 0.5674 \\
MHD1024 & 8 & 1.5923 & 1.0339 \\
Mixing & 1 & 0.1595 & 0.1086 \\
Mixing & 4 & 0.4983 & 0.2981 \\
Mixing & 8 & 1.0909 & 0.5358 \\ 
boundary layer & None & 3.7822 & 2.1542 \\ \bottomrule
\end{tabular}
\end{table}

\begin{tikzpicture}
    \begin{axis}[
        ybar,
        bar width=.2cm,
        width=8cm,
        height=6cm,
        enlarge x limits={abs=0.8cm},
        legend style={at={(0.5,-0.15)}, anchor=north, legend columns=-1},
        ylabel={Velocity (Pixel/Frame)},
        ylabel near ticks,
        xlabel={Scaling},
        symbolic x coords={1, 4, 8},
        xtick=data,
        ymin=0,
        grid=major,
        ymajorgrids=true,
        tick label style={font=\small},
        label style={font=\small},
        legend style={font=\small},
        title={}
    ]

    \addplot+[
        error bars/.cd,
        y dir=both, y explicit
    ] coordinates {
        (1,0.8939) +- (0,0.0843)
        (4,3.5547) +- (0,0.2675)
        (8,7.0262) +- (0,0.5067)
    };
    \addlegendentry{Channel}

    \addplot+[
        error bars/.cd,
        y dir=both, y explicit
    ] coordinates {
        (1,1.0388) +- (0,0.4291)
        (4,4.5454) +- (0,1.8380)
        (8,8.6357) +- (0,4.2400)
    };
    \addlegendentry{Isotropic1024}

    \addplot+[
        error bars/.cd,
        y dir=both, y explicit
    ] coordinates {
        (1,0.2393) +- (0,0.1406)
        (4,1.0064) +- (0,0.5674)
        (8,1.5923) +- (0,1.0339)
    };
    \addlegendentry{MHD1024}

    \addplot+[
        error bars/.cd,
        y dir=both, y explicit
    ] coordinates {
        (1,0.1595) +- (0,0.1086)
        (4,0.4983) +- (0,0.2981)
        (8,1.0909) +- (0,0.5358)
    };
    \addlegendentry{Mixing}

    \end{axis}
\end{tikzpicture}

\begin{figure}[htb]
    \centering
    \begin{subfigure}[b]{0.35\textwidth}
        \centering
        \includegraphics[width=\textwidth]{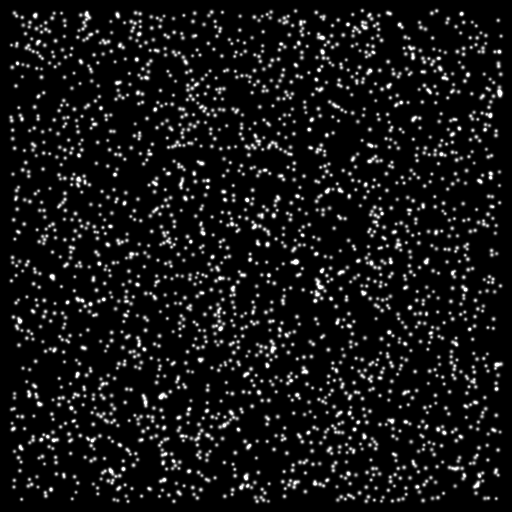}
        \caption{Density=0.01}
        \label{Density=0.01}
    \end{subfigure}
    \hfill
    \begin{subfigure}[b]{0.35\textwidth}
        \centering
        \includegraphics[width=\textwidth]{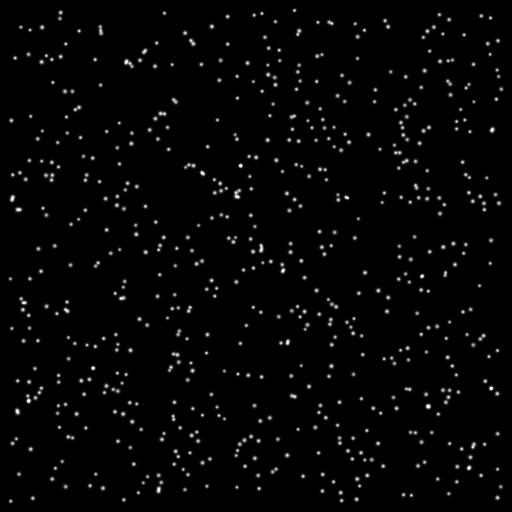}
        \caption{Density=0.025}
        \label{Density=0.025}
    \end{subfigure}
    \hfill
    \begin{subfigure}[b]{0.35\textwidth}
        \centering
        \includegraphics[width=\textwidth]{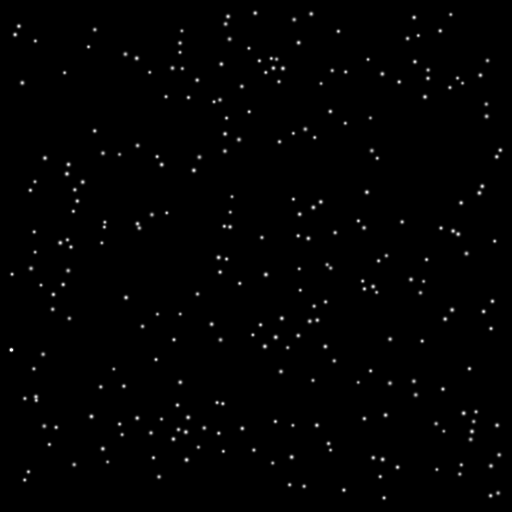}
        \caption{Density=0.001}
        \label{Density=0.001}
    \end{subfigure}
    \caption{Three consecutive images.}
    \label{density}
\end{figure}

The images of different particle densities are shown in Figure \ref{density}. These three particle densities represent different PIV particle concentrations: dense, moderate, and sparse. It's worth noting that at a particle density of 0.01, the distinction between PIV and optical flow is less pronounced. With a sufficiently dense particle distribution, predicting pixel motion can approximate flow field prediction. However, as particle density decreases, large regions of the image contain only black background. From an optical flow perspective, these regions would have zero motion. In contrast, within the PIV framework, these regions still possess a flow field, which must be inferred from the motion of surrounding particles.

\section{Multi Cost Volume PIV (MC-PIV)}
\begin{figure*}[htbp]
    \centering
    \includegraphics[width=1\textwidth]{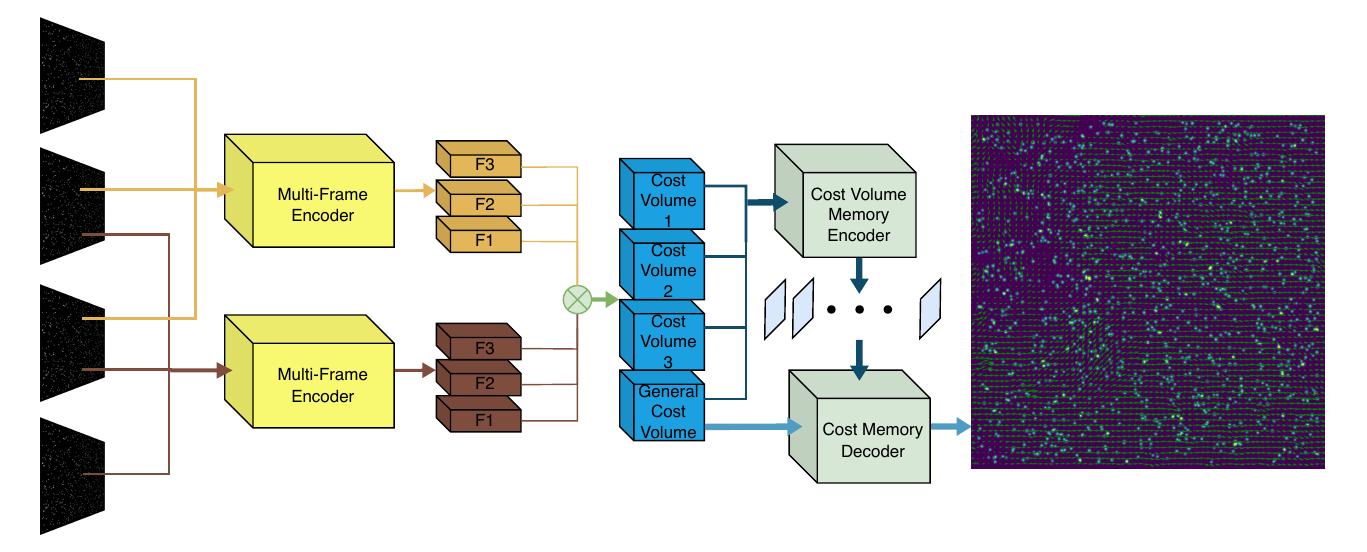}   
    \caption{The Architecture of the Multiframe-former}    
    \label{model} 
\end{figure*}

We propose a novel multi-frame and multi-cost-volume PIV model (MC-PIV), engineered upon the FlowFormer architecture \cite{flowformer}. As illustrated in Figure~\ref{model}, MC-PIV processes four sequential image frames as input to predict the fluid flow field between the central two frames ($t_2$ and $t_3$). The key innovation lies in its sophisticated feature extraction and its explicit use of multiple, distinct cost volumes to capture rich spatio-temporal information, significantly enhancing flow prediction accuracy, particularly for PIV data.

\subsection{Multi-Frame Block (MF Block):}
\label{subsec:mf_block}

\begin{figure*}[htbp]
    \centering
    \includegraphics[width=1\textwidth]{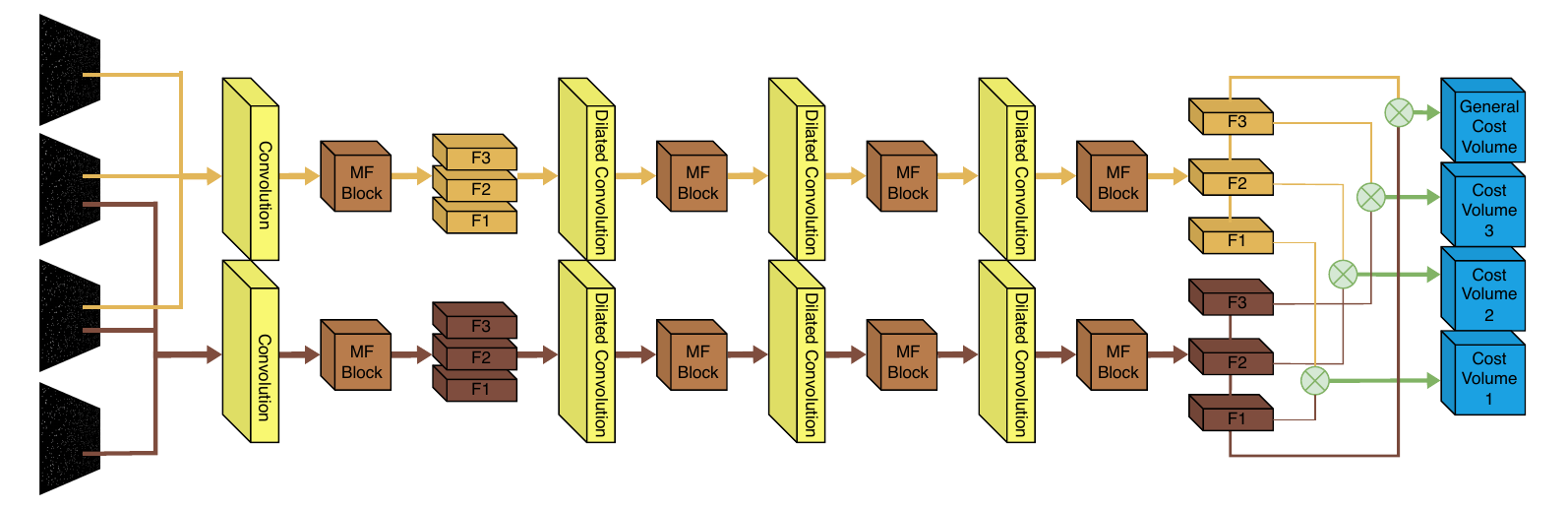}   
    \caption{The Feature Extraction Part}    
    \label{Encoder} 
\end{figure*}
The foundation of our temporal feature extraction is the Multi-Frame Block (MF Block), depicted in Figure~\ref{Encoder}. This block is designed to effectively leverage both inter-frame and intra-frame relationships using attention mechanisms.

The core component for capturing temporal dependencies in our architecture is the Multi-Frame Block (MF Block), illustrated in Figure~\ref{Encoder}. This block processes sequences of three consecutive frames to extract enriched spatio-temporal features by selectively employing attention mechanisms.

Let the input to the MF Block be three consecutive frames $I(t_1), I(t_2), I(t_3)$. We designate the central frame, $I(t_2)$, as the \textit{reference frame} for attention computation within this block instance. Let $S(t_i)$ represent the initial feature representation of frame $I(t_i)$ after preliminary shared convolutional layers. The MF Block then applies attention as follows:
\begin{itemize}
    \item \textbf{Self-Attention (SA) on Reference Frame:} The features of the reference frame, $S(t_2)$, undergo self-attention ($\text{SA}$) to capture complex spatial relationships and context within that specific time step.
    \item \textbf{Cross-Attention (CA) on Adjacent Frames:} The features of the adjacent frames, $S(t_1)$ and $S(t_3)$, undergo cross-attention ($\text{CA}$) with respect to the reference frame. Specifically, $S(t_1)$ and $S(t_3)$ provide the keys and values, while the reference frame features $S(t_2)$ provide the queries. This mechanism allows the block to correlate information from adjacent time steps ($t_1, t_3$) with the context of the central reference frame ($t_2$).
\end{itemize}
Following the attention operations, dedicated processing pathways ($E_1, E_2, E_3$), typically involving further convolutions or transformations, produce the final enhanced feature maps $F(t_1), F(t_2), F(t_3)$ for the three input time steps:
\begin{align*}
    F(t_2) &= E_1(\text{SA}(S(t_2))) \\
    F(t_1) &= E_2(\text{CA}(S(t_1), S(t_2))) \\
    F(t_3) &= E_3(\text{CA}(S(t_3), S(t_2)))
\end{align*}
These output features, $F(t_i)$, encapsulate richer temporal context compared to single-frame processing and serve as the basis for constructing multiple cost volumes in the subsequent stage.
\subsection{Two-Stream Architecture and Multi-Cost-Volume Integration}
\label{subsec:two_stream_integration}

To comprehensively capture the motion between the target frames ($t_2$ and $t_3$) while leveraging extended temporal context, our MC-PIV model employs a two-stream feature extraction architecture, explicitly designed for the generation and integration of \textbf{four distinct cost volumes}.

\pmb{Parallel Feature Extraction Streams:} The architecture comprises two parallel streams processing overlapping temporal windows, each utilizing the MF Blocks described above:
\begin{itemize}
    \item \textbf{Stream 1:} Processes input frames $I(t_1), I(t_2), I(t_3)$, using $I(t_2)$ as the reference frame within its MF Block. Outputs enhanced features $F_1(t_1), F_1(t_2), F_1(t_3)$.
    \item \textbf{Stream 2:} Processes input frames $I(t_2), I(t_3), I(t_4)$, using $I(t_3)$ as the reference frame within its MF Block. Outputs enhanced features $F_2(t_2), F_2(t_3), F_2(t_4)$.
\end{itemize}
Input particle images are initially processed by shared large-kernel convolution layers to densify sparse PIV features, improving the efficacy of subsequent attention mechanisms. Dilated convolutions are strategically employed within the feature extraction pathways to manage computational cost while maintaining receptive field size.

\pmb{Multiple Cost Volume Generation:} Leveraging the temporally enriched features from both parallel streams, we construct a total of \textbf{four distinct cost volumes} to provide diverse perspectives on the motion dynamics, comprising three local volumes and one comprehensive general volume:

\begin{itemize}
    \item \textbf{Local Cost Volumes (CV\textsubscript{L1}, CV\textsubscript{L2}, CV\textsubscript{L3}):} Three volumes capture fine-grained motion correlations using specific cross-stream feature pairings to detail the flow evolution into, during, and out of the target prediction interval ($t_2 \rightarrow t_3$):
        \begin{itemize}
            \item CV\textsubscript{L1} correlates $F_1(t_1)$ and $F_2(t_2)$ (pre-interval vs. interval start).
            \item CV\textsubscript{L2} correlates $F_1(t_2)$ and $F_2(t_3)$ (during the target interval).
            \item CV\textsubscript{L3} correlates $F_1(t_3)$ and $F_2(t_4)$ (interval end vs. post-interval).
        \end{itemize}

    \item \textbf{General Cost Volume (CV\textsubscript{G}):} One comprehensive volume capturing the overall relationship between the two temporal windows. It is computed by performing a \textbf{correlation} operation between the aggregated features from each stream. Specifically, it correlates the set of features from Stream 1, $\{F_1(t_1), F_1(t_2), F_1(t_3)\}$, with the set of features from Stream 2, $\{F_2(t_2), F_2(t_3), F_2(t_4)\}$. This captures the global correspondence and temporal shift between the contexts represented by the two streams.

\end{itemize}
These four cost volumes (CV\textsubscript{L1}, CV\textsubscript{L2}, CV\textsubscript{L3}, CV\textsubscript{G}) collectively form the rich multi-perspective input for the subsequent flow prediction stage.

\textbf{Flow Prediction via Adapted FlowFormer:} We adapt the cost processing modules of the FlowFormer \cite{flowformer} architecture to effectively utilize this multi-cost-volume information for predicting the optical flow between $I(t_2)$ and $I(t_3)$:
\begin{itemize}
    \item \textbf{Cost Memory Encoding:} The \textbf{four distinct cost volumes} (CV\textsubscript{L1}, CV\textsubscript{L2}, CV\textsubscript{G1}, CV\textsubscript{G2}) are simultaneously input into an adapted Cost Memory Encoder. This encoder module is responsible for fusing the information from these diverse cost representations into a single, compact, yet comprehensive cost memory.
    \item \textbf{Cost Memory Decoding:} This unified cost memory, potentially augmented by one of the general cost volumes serving as an explicit cost map, is then processed by an adapted Cost Memory Decoder. Through iterative refinement steps querying the rich cost memory, the decoder estimates the final high-resolution flow field.
\end{itemize}
By explicitly constructing and integrating information from these \textbf{multiple cost volumes}, derived from different temporal contexts and feature granularities, MC-PIV achieves a more robust and accurate representation of complex particle motion dynamics inherent in PIV data.

\section{Benchmark:}

\subsection{Introduction To The Model:}

\textbf{FlowFormer:} FlowFormer\cite{flowformer}, a transformer-based architecture for optical flow, encodes a 4D cost volume into a compact cost memory via an encoder with alternate-group transformer layers and decodes it through a recurrent transformer decoder with dynamic positional cost queries.
    
\textbf{LiteFlowNet:} LiteFlowNet\cite{LiteFlowNet}, a lightweight CNN for optical flow estimation, uses a pyramidal feature extractor, cascaded flow inference with feature warping, and a novel flow regularization layer based on feature-driven local convolution.

\textbf{FlowNet:} FlowNet\cite{flownet} proposes two convolutional neural network architectures for optical flow estimation: a "simple" architecture (FlowNetS) that processes concatenated input images, and a "correlation" architecture (FlowNetC) that uses separate processing streams for each image and incorporates a correlation layer for explicit feature matching before combining them.

\textbf{RAFT:}RAFT\cite{RAFT} employs a recurrent update operator that iteratively refines a high-resolution flow field by leveraging per-pixel features and lookups on multi-scale 4D all-pairs correlation volumes.

\textbf{SEA-RAFT:}SEA-RAFT\cite{sea-raft} is a modified RAFT architecture that regresses an initial flow, uses a Mixture-of-Laplace loss, and incorporates rigid-motion pre-training while simplifying the backbone and recurrent unit with standard ResNets and ConvNext blocks.

\textbf{GMA:}GMA\cite{GMA} uses a transformer-based global motion aggregation module to aggregate motion features based on self-similarities learned from the reference frame, augmenting a RAFT architecture for improved optical flow estimation in occluded regions.

\textbf{GMFlowNet:}GMFlowNet\cite{gmflownet} incorporates a global matching step, computed via argmax on a 4D cost volume, before a RAFT-based iterative optimization module, and uses patch-based overlapping attention to enhance feature extraction.

\textbf{StreamFlow}: StreamFlow\cite{StreamFlow} proposes a streamlined in-batch multi-frame (SIM) pipeline for efficient optical flow estimation in video sequences, eliminating redundant recursive computations. It introduces an Integrative Spatio-temporal Coherence (ISC) module in the encoder, which integrates temporally contiguous input embeddings for parameter-efficient spatio-temporal modeling, and a Global Temporal Regressor (GTR) in the decoder that utilizes super convolution kernels and a lightweight temporal transformer to exploit temporal cues for flow refinement.


\begin{table*}[ht]
\centering
\begin{tabular}{l r@{\;}r r@{\;}r r@{\;}r r@{\;}r r@{\;}r r@{\;}r}
\toprule
model & \multicolumn{2}{c}{MHD} & \multicolumn{2}{c}{Isotropic} & \multicolumn{2}{c}{Mixing} & \multicolumn{2}{c}{Channel} & \multicolumn{2}{c}{\makebox[1cm]{Boundary Layer}} & \multicolumn{2}{c}{All Dataset} \\
 & EPE & NEPE & EPE & NEPE & EPE & NEPE & EPE & NEPE & EPE & NEPE & EPE & NEPE \\
\midrule

FlowNetc & 1.316 & 1.938 & 5.873 & 0.996 & 1.063 & 2.580 & 3.410 & 0.938 & 1.199 & 0.538 & 2.784 & 1.530 \\
FlowNets & 0.623 & 0.923 & 3.112 & 0.519 & 0.442 & 1.007 & 0.479 & 0.174 & 0.308 & 0.187 & 1.098 & 0.620 \\
GMA & 0.645 & 0.875 & 2.324 & 0.464 & 0.462 & 0.961 & 0.391 & 0.113 & 0.242 & 0.175 & 0.901 & 0.570 \\
GMFlow & 1.046 & 1.109 & 9.506 & 1.468 & 0.611 & 1.039 & 4.688 & 1.320 & 3.993 & 1.142 & 3.965 & 1.227 \\
GMFlowNet & 0.922 & 1.082 & 3.924 & 0.674 & 0.517 & 1.020 & 1.407 & 0.409 & 0.625 & 0.335 & 1.610 & 0.761 \\
LiteFlowNet & 0.918 & 1.155 & 3.253 & 0.645 & 0.534 & 1.133 & 0.481 & 0.171 & 0.977 & 0.537 & 1.272 & 0.758 \\
LiteFlowNet3s & 0.992 & 1.249 & 3.415 & 0.684 & 0.605 & 1.268 & 0.661 & 0.233 & 1.003 & 0.600 & 1.386 & 0.839 \\
RAFT & 0.766 & 1.004 & 2.646 & 0.523 & 0.482 & 1.008 & 0.435 & 0.127 & 0.226 & 0.099 & 1.016 & 0.622 \\
SeaRAFT & 1.076 & 1.391 & 4.880 & 0.885 & 0.641 & 1.412 & 1.978 & 0.520 & 2.521 & 1.096 & 2.173 & 1.055 \\
StreamFlow & 0.614 & 0.770 & 4.903 & 0.653 & 0.490 & 1.003 & 0.369 & 0.105 & \pmb{0.061} & \pmb{0.026} & 1.476 & 0.586 \\
FlowFormer & 0.483 & 0.721 & $\pmb{1.905}$ & $\pmb{0.369}$ & 0.352 & 0.760 & 0.328 & 0.102 & 0.094 & 0.041 & $\pmb{0.715}$ & 0.454 \\
\midrule
$\pmb{\text{MCFormer}}$ & $\pmb{0.241}$ & $\pmb{0.394}$ & 3.759 & 0.418 & $\pmb{0.158}$ & $\pmb{0.385}$ & $\pmb{0.195}$ & $\pmb{0.065}$ & 0.149 & 0.055 & 1.016 & $\pmb{0.295}$ \\
\midrule
Average & 0.803 & 1.051 & 4.125 & 0.692 & 0.530 & 1.131 & 1.235 & 0.357 & 0.950 & 0.403 & 1.618 & 0.776 \\
\bottomrule
\end{tabular}
\caption{Average test EPE and Normalized EPE (NEPE) for different models across the five PIV datasets. Best results for each metric per dataset are in bold. Best overall results are also bolded.}
\label{tab:average_results}
\end{table*}
\setlength{\tabcolsep}{6pt} 
\subsection{The Experiments Set:}

We employed the endpoint error (EPE) loss as both the training objective and the evaluation metric for assessing model performance. The EPE loss is calculated as equation~\eqref{eq:epe_loss}. 
\begin{equation}
EPE = \frac{1}{N} \sum_{i=1}^{N} ||\mathbf{u}_i^{pred} - \mathbf{u}_i^{gt}||_2, \label{eq:epe_loss}
\end{equation}
where \(N\) is the number of pixels (or points) in the flow field, \(\mathbf{u}_i^{pred}\) is the predicted flow vector at pixel \(i\), \(\mathbf{u}_i^{gt}\) is the ground truth flow vector at pixel \(i\), and \(|| \cdot ||_2\) denotes the L2 norm (Euclidean distance).
It provides a direct measure of the pixel-wise difference between the predicted flow field and the ground truth. This metric is widely adopted in optical flow and PIV literature due to its clear physical interpretation and ease of computation.

In addition to the standard EPE loss, we introduce a metric, Normalize EPE (NEPE), to better assess prediction accuracy across varying flow velocities. NEPE is calculated as equation\eqref{eq:NEPE_loss}.

\begin{equation}
NEPE = \frac{1}{N} \sum_{i=1}^{N} \frac{||\mathbf{u}_i^{pred} - \mathbf{u}_i^{gt}||_2}{||\mathbf{u}_i^{gt}||_2 + \epsilon}, \label{eq:NEPE_loss}
\end{equation}
where the variables are the same as defined for the EPE loss.
 This normalization provides a percentage representation of the error relative to the true flow magnitude, allowing for a more meaningful comparison of model performance across different flow speeds. Specifically, a lower NEPE value indicates a smaller error relative to the actual flow velocity, signifying better performance, particularly in regions with high flow speeds. This metric addresses the potential issue where larger absolute EPE values might simply reflect higher overall flow magnitudes rather than necessarily poorer model performance.

To ensure a fair comparison between different models, we maintained consistent training settings across all experiments. Specifically, we utilized the Adam optimizer. The learning rate was initially set to 1e-4. Each model was trained for a maximum of 100 epochs with a batch size of 1. Early stopping was applied when the EPE loss on the held-out validation set failed to decrease for five consecutive epochs. This early stopping strategy helps prevent overfitting and ensures that the comparisons are made at a similar level of convergence for each model.

\subsection{result and analysis:} 

The experimental results, summarized in Table~\ref{tab:average_results}, demonstrate the competitive performance of our proposed MCFormer model against established optical flow methods adapted for PIV, and includes the newly evaluated multi-frame model, StreamFlow. Our model (MCFormer) achieves the best overall performance in terms of Normalized Endpoint Error (NEPE), with an average score of \pmb{0.295} across all datasets. This indicates a strong capability to maintain accuracy relative to the true flow magnitude, which is particularly valuable in PIV where velocity ranges can vary significantly. FlowFormer retains the best overall absolute performance (EPE) with a score of \pmb{0.715}.

Specifically, MCFormer significantly outperforms all baseline models on the Channel, MHD, and Mixing datasets, securing the lowest EPE and NEPE scores in these categories. This highlights its effectiveness in handling flows with varying degrees of structure and turbulence levels, especially at lower to moderate average speeds (as indicated in Table~\ref{velTable}). In contrast, StreamFlow, despite also being a multi-frame approach leveraging temporal information, shows significantly weaker performance on these three datasets, underscoring that multi-frame capabilities alone do not guarantee success across all PIV conditions.

\begin{figure}[htb]
    \centering
    \subfloat{\includegraphics[width=0.45\textwidth]{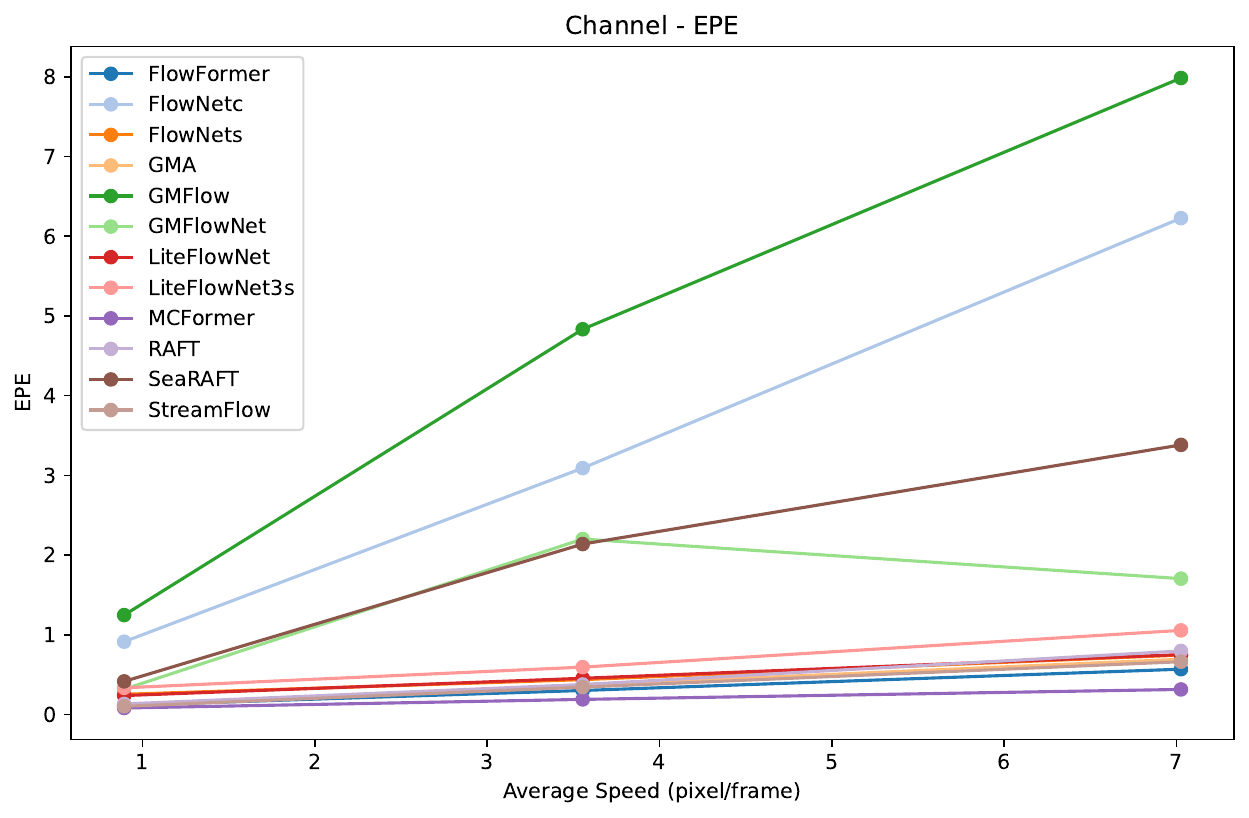}}
    \hfill
    \subfloat{\includegraphics[width=0.45\textwidth]{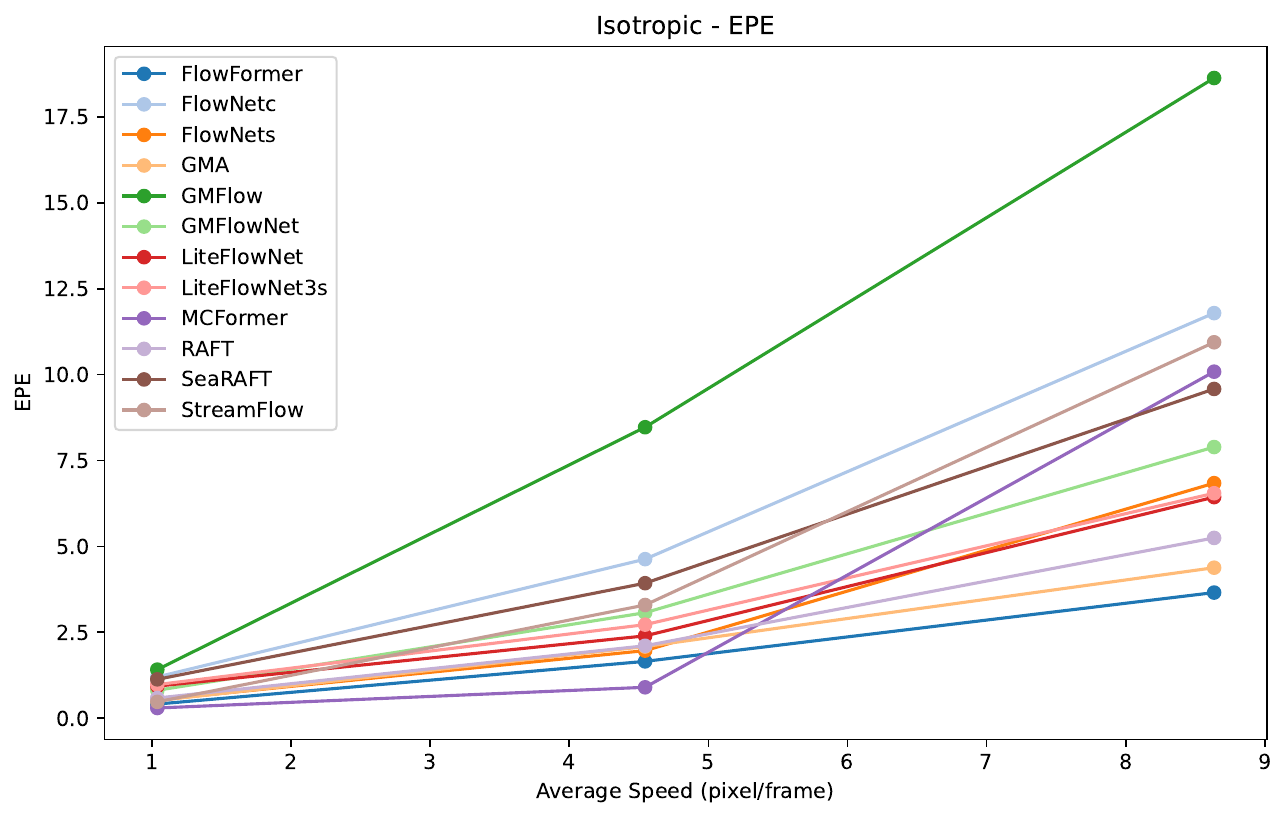}}\\
    \subfloat{\includegraphics[width=0.45\textwidth]{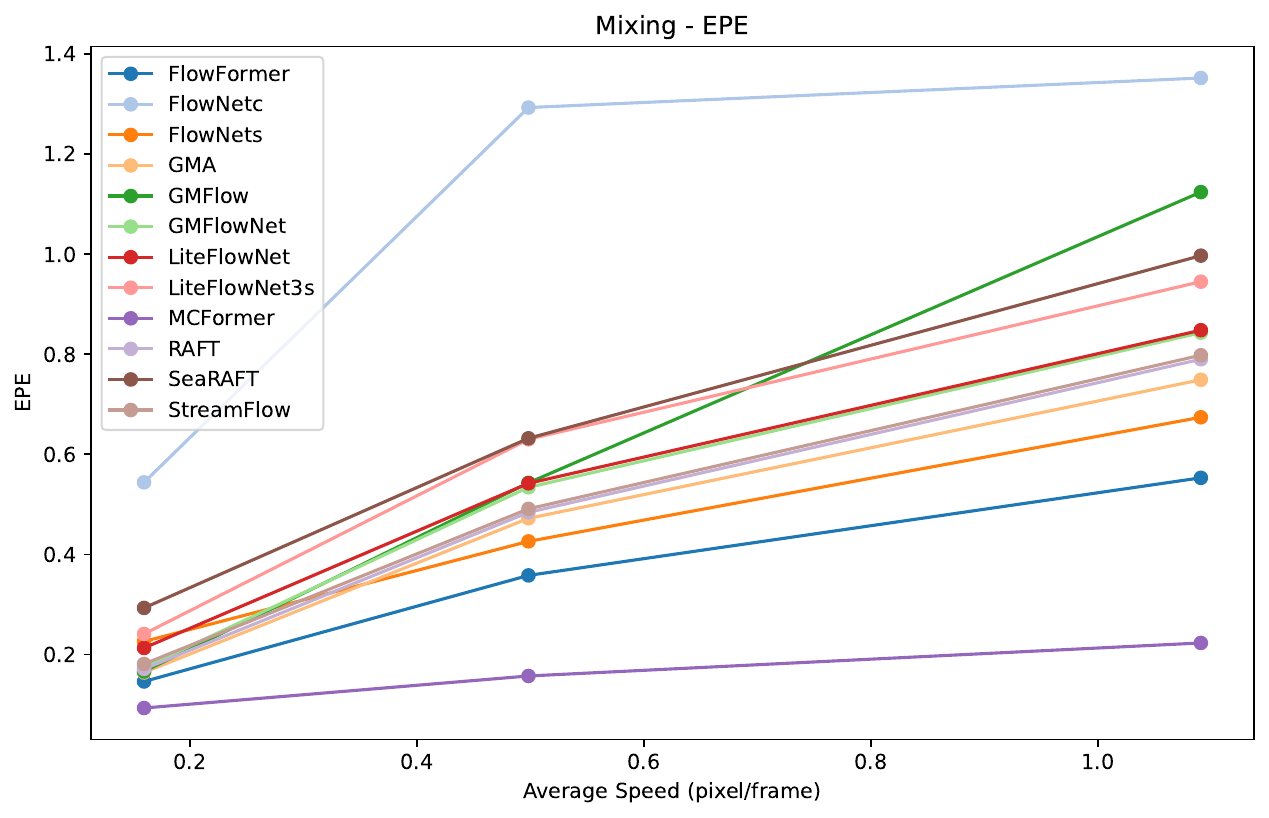}}
    \hfill
    \subfloat{\includegraphics[width=0.45\textwidth]{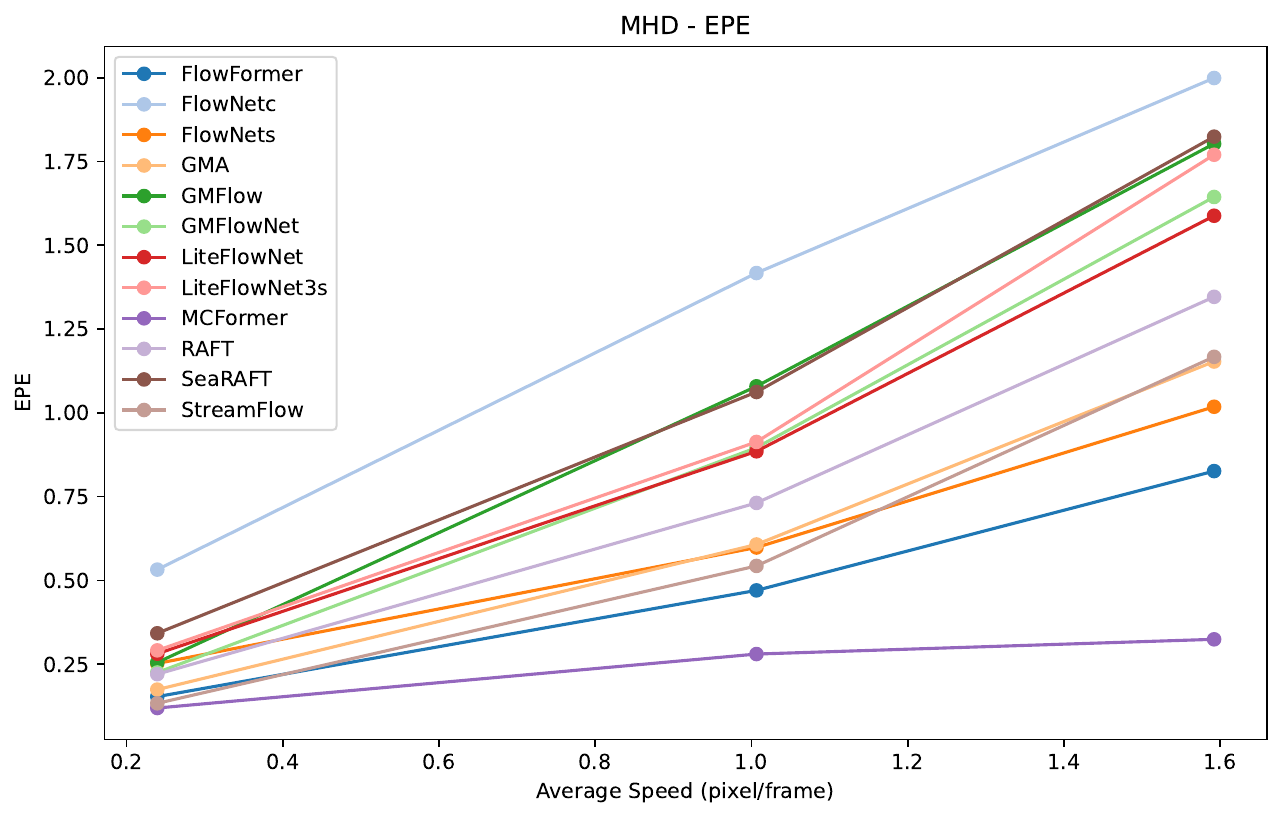}}
    \caption{Relationship between EPE and flow field velocity}
    \label{epeVsSpeed}
\end{figure}

However, FlowFormer achieves the best overall EPE with a score of \pmb{0.715}, demonstrating superior performance in terms of absolute error on average. FlowFormer also leads on the Isotropic and Boundary Layer datasets. A key factor contributing to MCFormer's higher overall EPE is its performance on the Isotropic dataset, particularly at the highest flow speed scaling (8x speed factor). As depicted in Figure~\ref{epeVsSpeed} (top-right panel), MCFormer's EPE increases substantially in this challenging high-speed, highly turbulent regime, negatively impacting its overall average EPE. This suggests a potential sensitivity of MCFormer to extremely complex, high-velocity turbulent flows compared to FlowFormer. On the Boundary Layer dataset, MCFormer performs competitively, yielding EPE and NEPE values only slightly higher than FlowFormer.

The performance difference between models optimized for standard optical flow benchmarks (like Sintel and KITTI, see Table~\ref{tab:optical_flow_benchmarks}) and their performance on our PIV datasets underscores the unique challenges of PIV. Models like GMFlowNet and GMA, which perform well on Sintel/KITTI, exhibit relatively high errors on the PIV data, suggesting that architectures excelling at dense optical flow do not directly translate to optimal performance on sparse particle imagery. This reinforces the need for specialized PIV approaches like MCFormer.

Analyzing performance across different dataset characteristics reveals further insights. The Isotropic dataset, characterized by high complexity and randomness, poses a significant challenge, resulting in higher average errors for most models compared to the more structured Channel flow, despite similar average flow speeds (Table~\ref{velTable}). The MHD and Mixing datasets generally yield lower absolute EPE values, likely due to their inherently lower flow speeds. However, considering the NEPE metric, the relative errors on MHD and Mixing can be higher than on Channel or Isotropic for some models, potentially due to the difficulty in accurately capturing smaller, more subtle motions prevalent in these slower flows. This highlights the importance of evaluating both EPE and NEPE for a comprehensive understanding.

\begin{table}[h]
\centering
\begin{tabular}{lcc}
\toprule
Model & Sintel Clean & KITTI 2015 (train) \\
\midrule
FlowFormer & 1.16 & 4.09 \\
FlowNetc & 7.28 & - \\
FlowNets & 7.42 & - \\
GMA & 1.38 & 4.69 \\
GMFlowNet & 1.39 & 4.24 \\
LiteFlownet & 4.54 & - \\
LiteFlownet3s & 3.03 & 7.22 \\
RAFT & 1.6 & 5.04 \\
Sea-RAFT(S) & 1.27 & 4.61 \\
StreamFlow & 1.04  & 4.24 \\
\bottomrule
\end{tabular}
\caption{Performance comparison on standard optical flow benchmarks.}
\label{tab:optical_flow_benchmarks}
\end{table}

\begin{figure}[htb]
    \centering
    \subfloat{\includegraphics[width=0.45\textwidth]{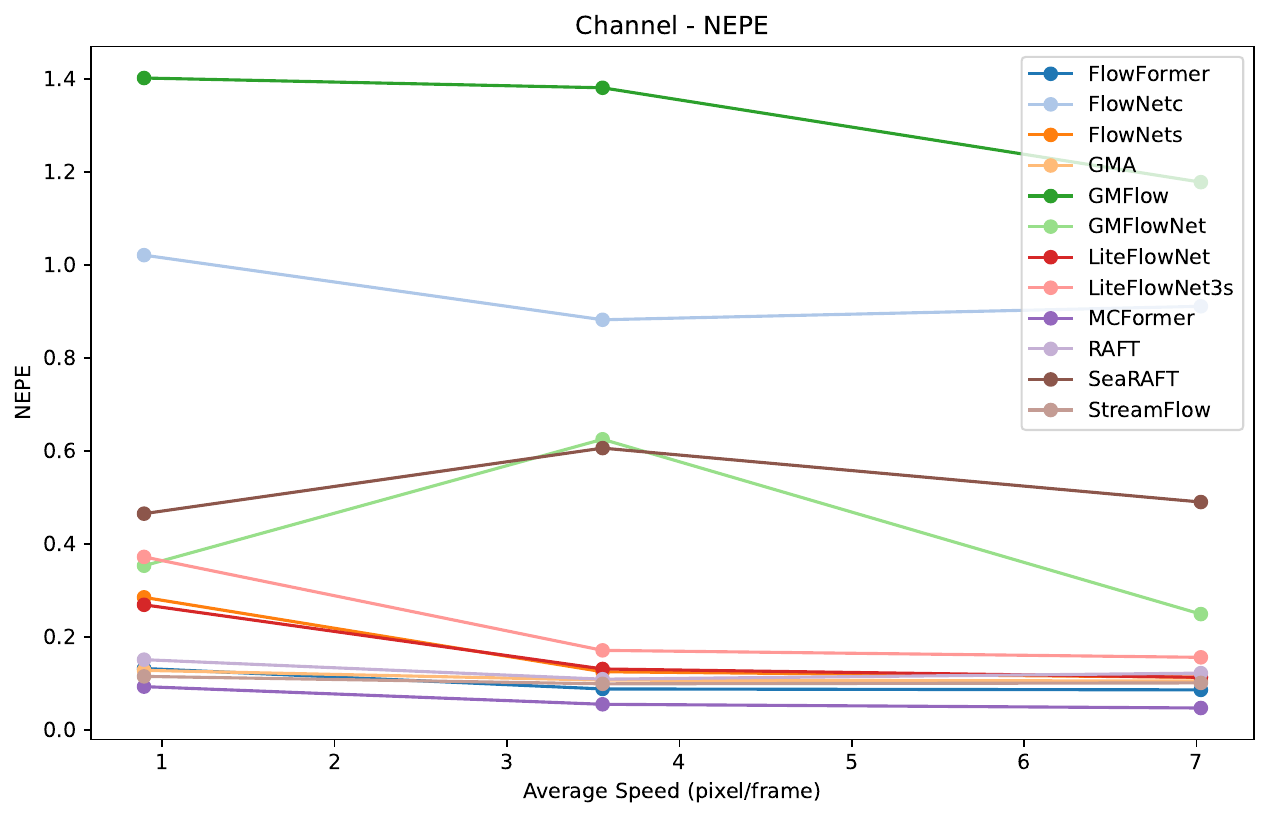}}
    \hfill
    \subfloat{\includegraphics[width=0.45\textwidth]{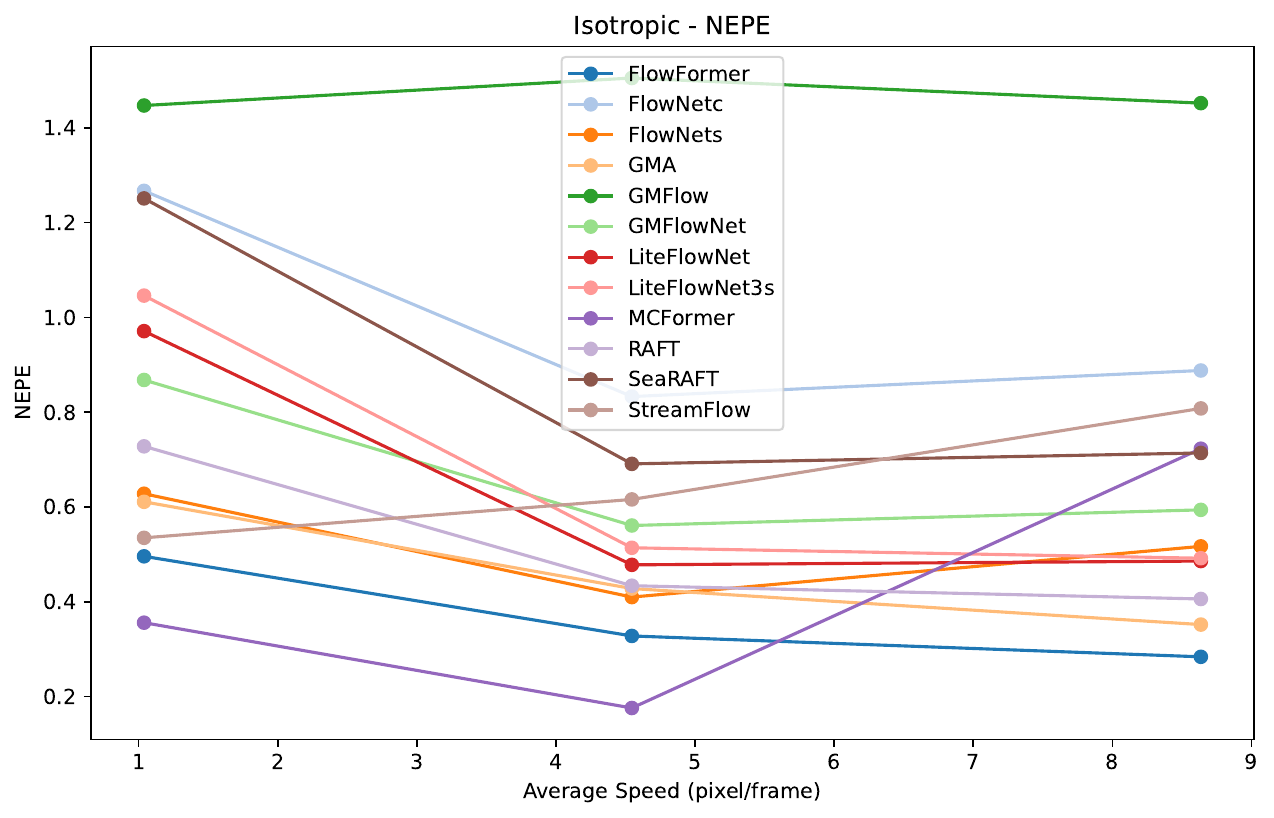}}\\
    \subfloat{\includegraphics[width=0.45\textwidth]{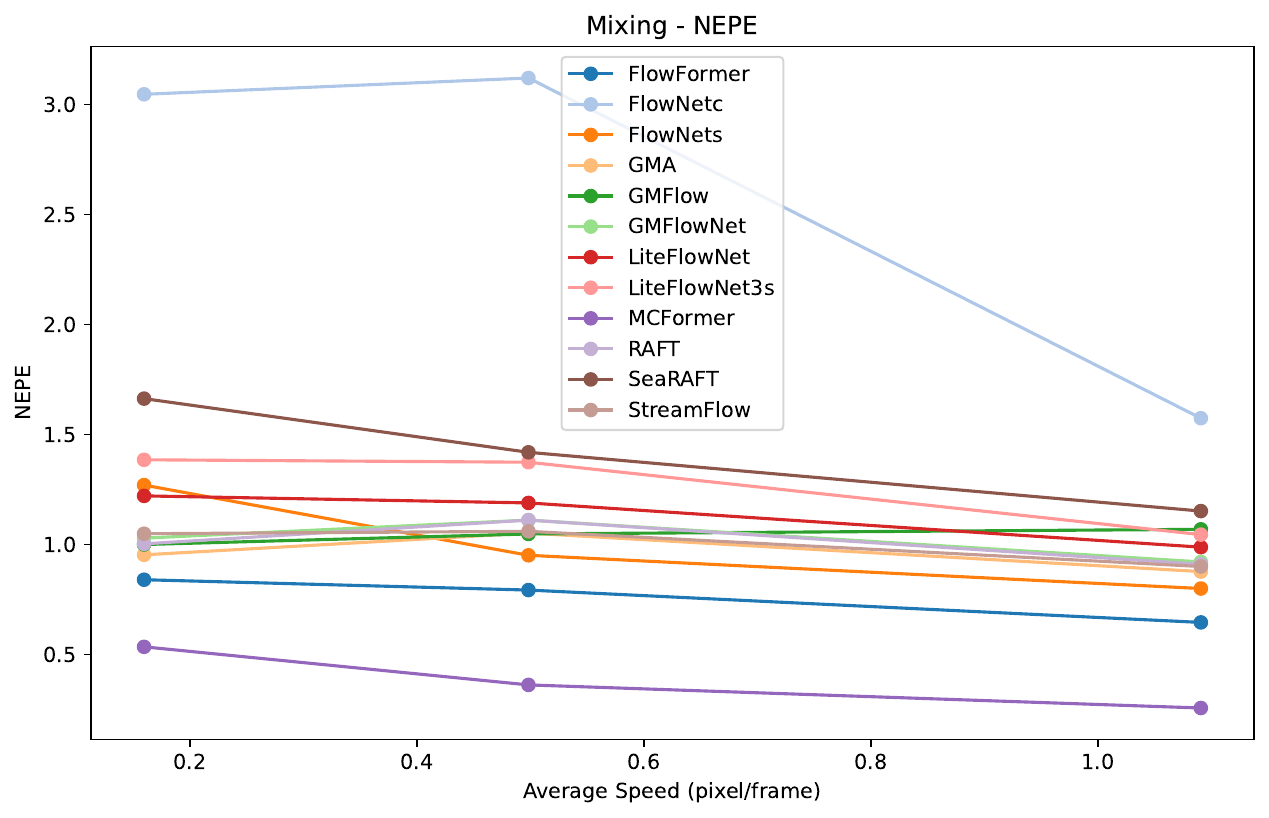}}
    \hfill
    \subfloat{\includegraphics[width=0.45\textwidth]{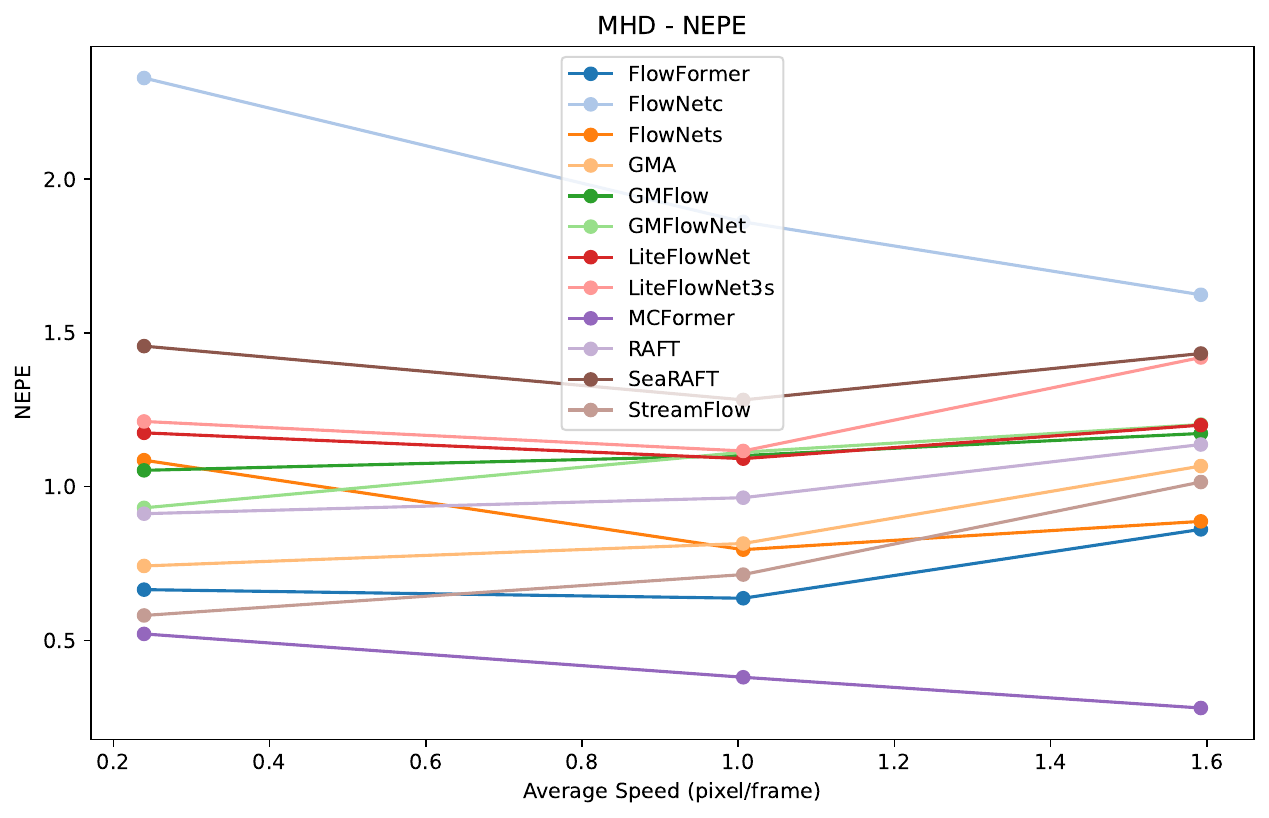}}
    \caption{Relationship between NEPE and flow field velocity}
    \label{nepeVsSpeed}
\end{figure}

This discrepancy underscores the inherent differences between Particle Image Velocimetry (PIV) tasks and traditional optical flow estimation tasks. The inferior performance of GMFlowNet and GMA on PIV suggests that architectures optimized for generic optical flow might not be directly transferable to the specific challenges presented by PIV.

\begin{figure}[htb]
    \centering
    \subfloat{\includegraphics[width=0.45\textwidth]{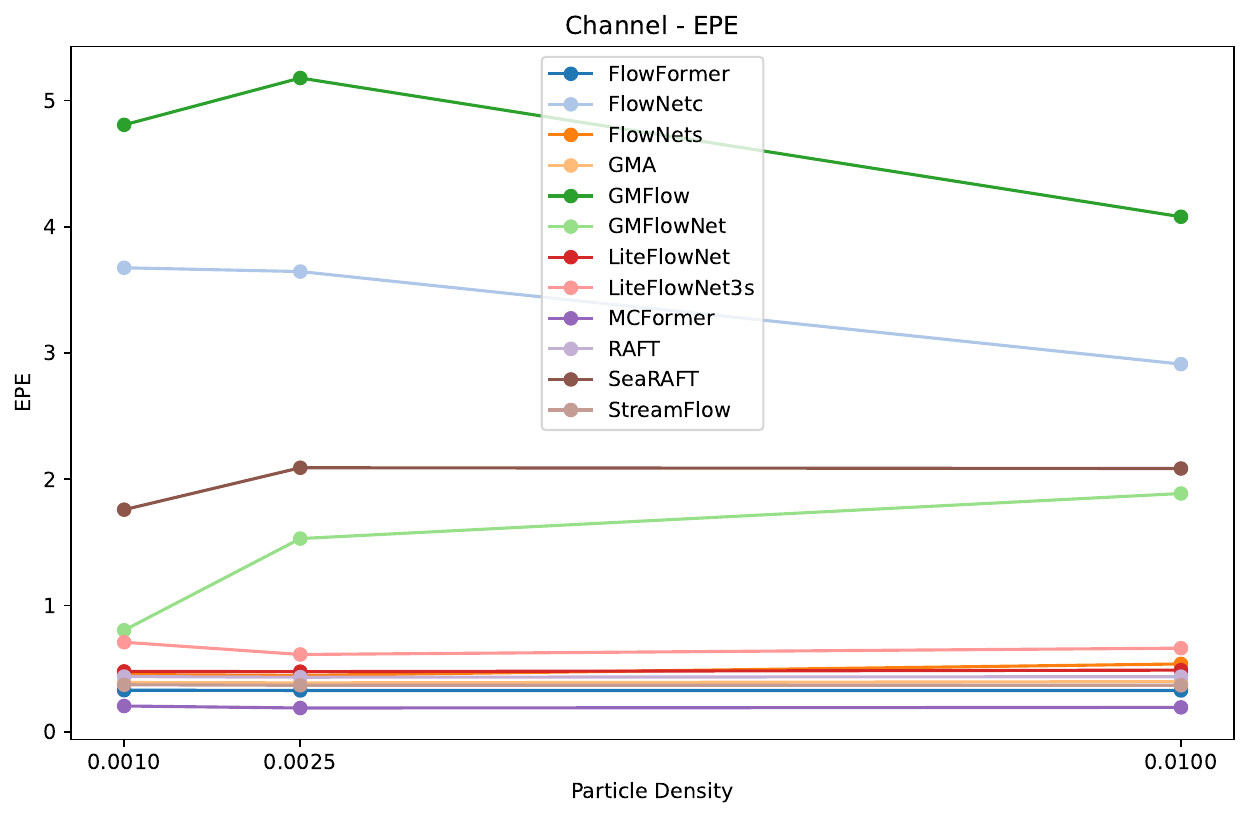}\label{fig:image1}}
    \hfill
    \subfloat{\includegraphics[width=0.45\textwidth]{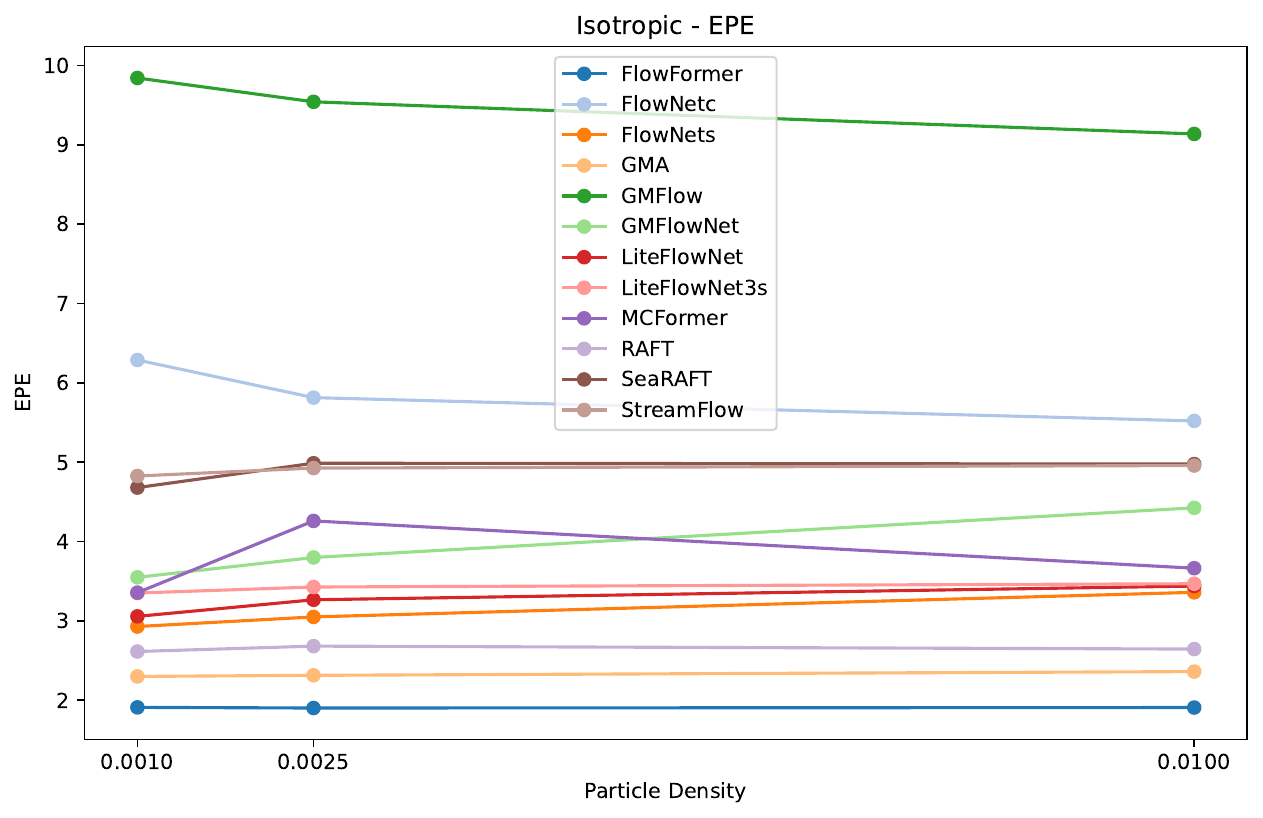}\label{fig:image2}}\\
    \subfloat{\includegraphics[width=0.45\textwidth]{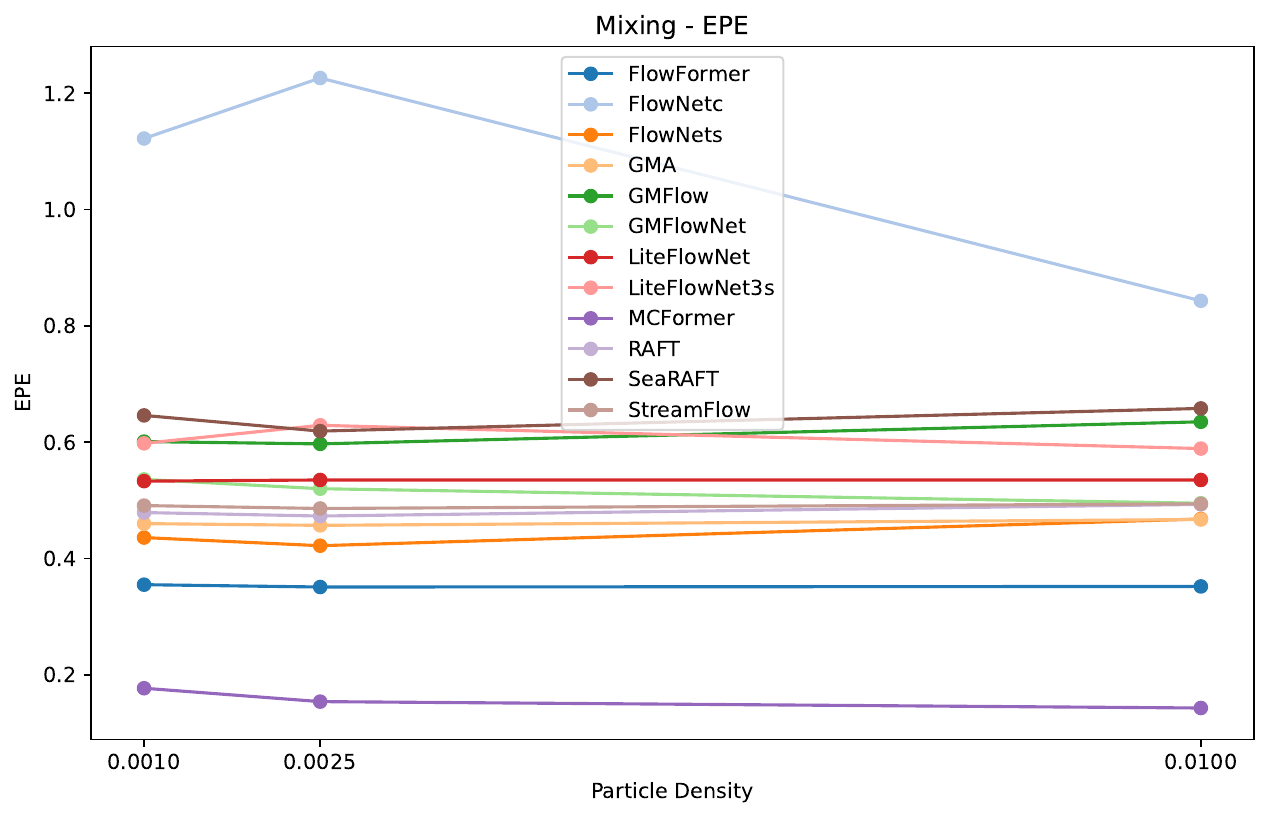}\label{fig:image3}}
    \hfill
    \subfloat{\includegraphics[width=0.45\textwidth]{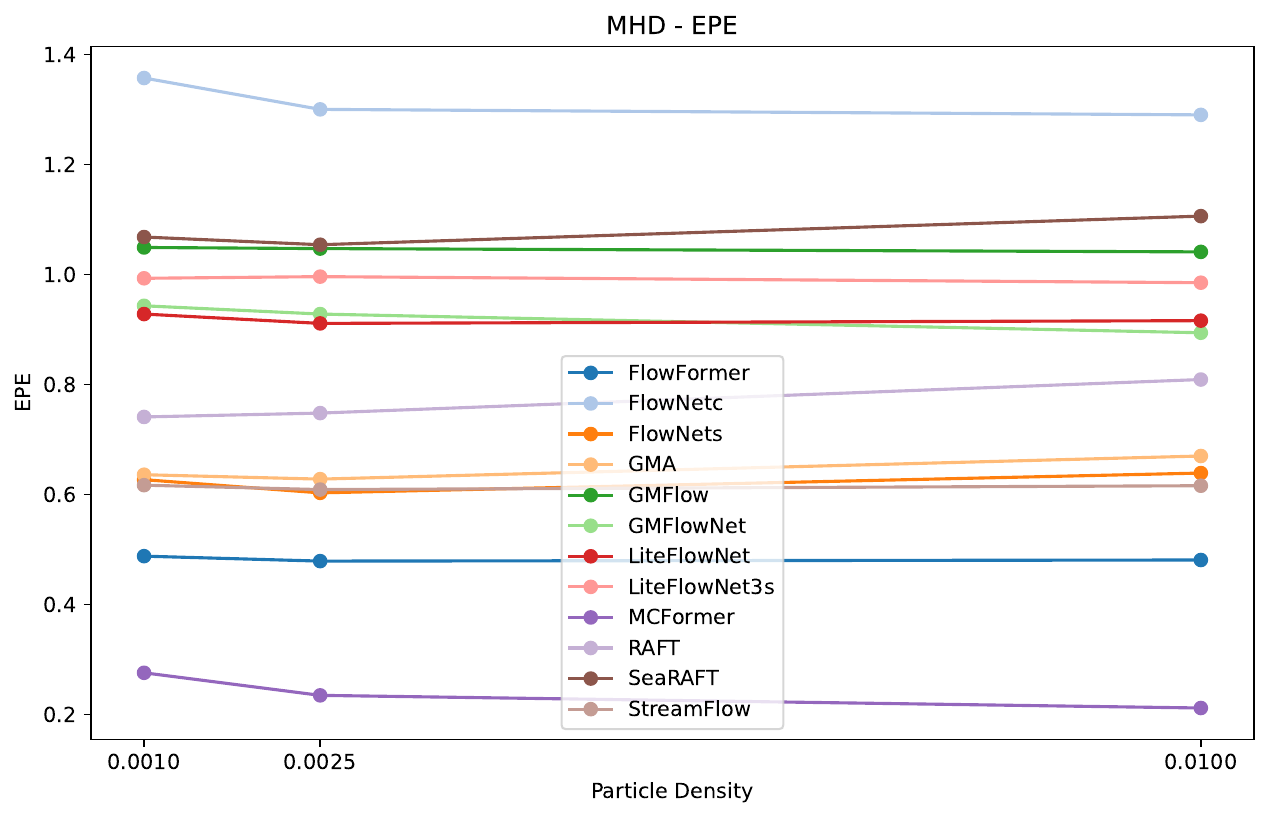}\label{fig:image4}}
    \caption{Relationship between EPE and particle density for different flow types}
    \label{epeVsDensity}
\end{figure}

\begin{figure}[htb]
    \centering
    \includegraphics[width=0.45\textwidth]{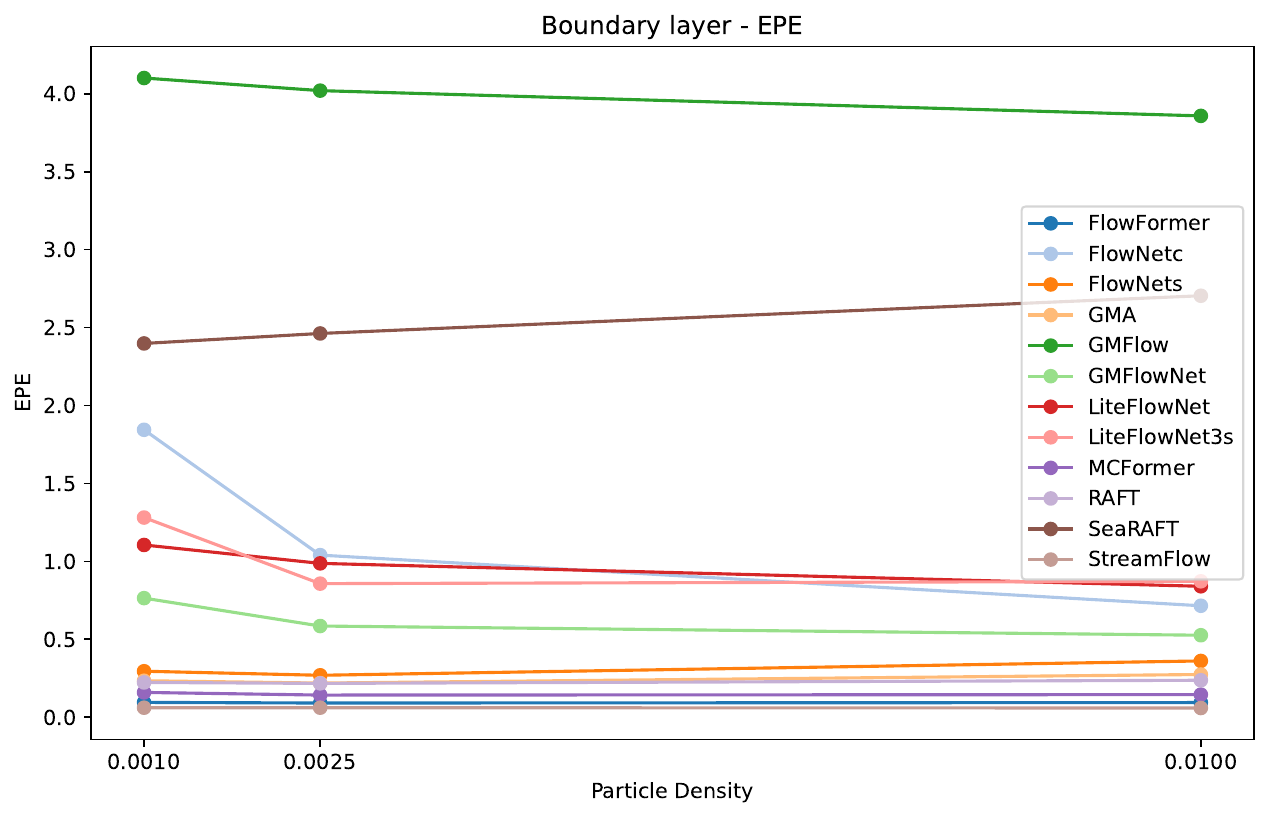}
    \caption{Relationship between EPE and particle density for boundary layer flow}
    \label{fig:boundary_layer_density}
\end{figure}

Analyzing the performance across different datasets reveals substantial variations in EPE and NEPE values. The "Isotropic" dataset presents the most significant challenge for all models, exhibiting the highest average EPE (4.054).Table~\ref{velTable} showing that the "Isotropic" and "Channel" datasets have similar flow speeds, the higher EPE and NEPE in "Isotropic" likely stems from the complexity and randomness of motion patterns inherent in this dataset. In contrast, the "Channel" dataset, despite having similar flow speeds to "Isotropic", results in the lowest average NEPE (0.379), likely due to more structured and predictable motion patterns. The "MHD" and "Mixing" datasets exhibit lower EPE values (0.821 and 0.533 respectively) compared to the "Isotropic" dataset. This observation, coupled with the knowledge that these datasets ("MHD" and "Mixing") inherently feature slower flow speeds, suggests that the magnitude of motion significantly influences the EPE metric. Lower flow speeds in "MHD" and "Mixing" likely contribute to smaller absolute errors. However, it's important to consider the NEPE values, which normalize the error by the magnitude of the flow. The higher NEPE values in "Mixing" and "MHD" comparing to "Isotropic" and "Channel" indicate that while absolute errors might be smaller , the normalize errors are larger, possibly due to the presence of smaller, more subtle motions that are challenging to accurately estimate. This analysis underscores the importance of considering both EPE and NEPE, as well as the inherent characteristics of each dataset, when evaluating the performance of optical flow estimation models.

Examining the relationship between error and flow speed (Figure~\ref{epeVsSpeed}), the EPE generally increases with velocity for most models and datasets, as expected. FlowFormer demonstrates competent performance, particularly securing the best results on the Isotropic dataset at the 8x scaling and on the Boundary Layer dataset. However, our proposed MCFormer exhibits superior performance across a vast majority of conditions. It achieves the \textbf{lowest EPE} on the Channel, MHD, and Mixing datasets across \textbf{all speed scales}, and also on the Isotropic dataset at the 1x and 4x speed scales. The only significant outlier is the Isotropic dataset at the 8x scaling, where MCFormer's EPE increases substantially, preventing it from achieving the best overall EPE despite its dominance elsewhere. On the Boundary Layer dataset, MCFormer consistently achieves the second-best EPE across all tested flow speeds, further highlighting its strong performance relative to most baselines.

Turning to relative error, Figure~\ref{nepeVsSpeed} depicts the NEPE variation with flow speed. The general trend of decreasing NEPE with increasing speed is observed for Channel, MHD, and Mixing datasets, potentially attributable to the EPE loss function prioritizing larger absolute errors during training. Critically, MCFormer achieves the \textbf{lowest NEPE} across the vast majority of dataset and speed combinations, culminating in the best overall NEPE score (Table~\ref{tab:average_results}). This underscores MCFormer's exceptional robustness in maintaining high accuracy relative to the true flow magnitude across diverse velocity regimes.

Finally, Figure~\ref{epeVsDensity} explores the impact of particle density on EPE. The results reveal a complex relationship, where accuracy often improves initially as density increases from sparse (0.001) to moderate (0.0025), but can then degrade for some models at the highest density (0.01), notably in the Channel dataset. This suggests potential challenges with particle overlap or feature extraction at high densities for certain architectures. In stark contrast, MCFormer demonstrates \textbf{remarkable robustness} to these variations, maintaining \textbf{consistently low EPE values across all tested densities}. This stability, unlike the significant fluctuations observed in models such as GMFlowNet, strongly highlights MCFormer's suitability and reliability for practical PIV applications where controlling particle seeding density perfectly is often challenging.

\section{Conclusion:}

This paper addressed critical limitations in deep learning-based PIV, primarily the absence of a comprehensive benchmark to evaluate how different optical flow algorithms perform on PIV tasks. We introduced two key contributions: 1) A novel, large-scale, diverse synthetic PIV benchmark dataset, and 2) MCFormer, a multi-frame deep learning model tailored for PIV. The dataset, derived from varied CFD sources, provides the much-needed standardized platform that facilitated the first extensive comparative analysis of deep learning PIV approaches presented herein. The MCFormer architecture effectively utilizes multi-frame temporal context and multiple cost volumes to handle sparse particle data.

Our experimental results, conducted on this new benchmark, demonstrate the strong performance of MCFormer, which achieved the best overall Normalized Endpoint Error (NEPE), indicating robustness across different flow magnitudes. More importantly, this benchmark evaluation provides the first quantitative comparison across a wide range of optical flow architectures (from FlowNetS to RAFT and FlowFormer) applied to PIV. The results reveal significant performance variations and confirm that models excelling on standard dense optical flow benchmarks (like Sintel/KITTI) do not necessarily translate effectively to the unique, often sparse, challenges of PIV, highlighting the need for specialized approaches and dedicated PIV benchmarks like the one we introduced.

While MCFormer shows significant promise, the evaluation also identified remaining challenges, particularly in maintaining consistently low relative errors (NEPE) in low-speed regimes and across all particle densities for all models. Future work should focus on developing algorithms with improved robustness to these factors, potentially through tailored attention mechanisms, physics-informed constraints, and further dataset expansion. Our benchmark provides the essential foundation for measuring progress in these future endeavors.

\section*{Acknowledgments}

We thank the Johns Hopkins Turbulence Database (JHTDB) for providing access to the high-quality CFD simulation data that made this work possible. We are grateful to the computational resources provided by Beijing University of Posts and Telecommunications Supercomputing Center. We also thank our colleagues for their valuable feedback and suggestions during the development of this work.

{\small
\bibliographystyle{ieeetr}
\bibliography{egbib}
}

\end{document}